\documentclass[runningheads]{llncs}

\usepackage{eccv}

\usepackage{eccvabbrv}

\usepackage{graphicx}
\usepackage{booktabs}
\usepackage{multirow}
\usepackage{newfloat}
\usepackage{listings}
\usepackage{xcolor}
\usepackage{booktabs}
\usepackage{multirow}
\usepackage{amsfonts}
\usepackage{makecell}
\usepackage{algorithm}
\usepackage{algorithmicx}
\usepackage{algpseudocode}
\usepackage{caption}
\usepackage{subcaption}
\usepackage{makecell}
\usepackage{bm}
\usepackage[accsupp]{axessibility}  % Improves PDF readability for those with disabilities.

\usepackage{hyperref}

\usepackage{orcidlink}

\begin{document}

% ---------------------------------------------------------------
% TODO REVIEW: Replace with your title
\title{Accelerating Image Generation with \\ Sub-Path Linear Approximation Model} 

% TODO REVIEW: If the paper title is too long for the running head, you can set
% an abbreviated paper title here. If not, comment out.
\titlerunning{SPLAM}

% TODO FINAL: Replace with your author list. 
% Include the authors' OCRID for the camera-ready version, if at all possible.
\author{Chen Xu\inst{1,2}$^\dag$\orcidlink{0009-0006-8534-0611} \and
Tianhui Song\inst{1,2}$^\dag$\orcidlink{0009-0007-4256-6407} \and 
Weixin Feng\inst{2}\orcidlink{0009-0002-7184-4115} \and
Xubin Li\inst{2}\orcidlink{0000-0002-1547-4052} \and \\
Tiezheng Ge\inst{2}\orcidlink{0000-0003-1381-2692} \and 
Bo Zheng\inst{2}\orcidlink{0000-0002-4037-6315} \and 
Limin Wang\inst{1,3,}$^*$\orcidlink{0000-0002-3674-7718}}

% TODO FINAL: Replace with an abbreviated list of authors.
\authorrunning{C. Xu et al.}
% First names are abbreviated in the running head.
% If there are more than two authors, 'et al.' is used.

% TODO FINAL: Replace with your institution list.
\institute{State Key Laboratory for Novel Software Technology, Nanjing University \and Alibaba Group \quad \quad \inst{3} Shanghai AI Lab\\ \url{https://subpath-linear-approx-model.github.io/}}

\maketitle

\let\thefootnote\relax\footnotetext{$^\dag$ Equal contributions. Interns at Alibaba Group. $^*$ Corresponding author.}

\begin{abstract}
  Diffusion models have significantly advanced the state of the art in image, audio, and video generation tasks. However, their applications in practical scenarios are hindered by slow inference speed. Drawing inspiration from the consistency models, we propose the \textbf{S}ub-\textbf{P}ath \textbf{L}inear \textbf{A}pproximation \textbf{M}odel (SPLAM), which can accelerate diffusion models while maintaining high-quality image generation. SPLAM treats the PF-ODE trajectory as a series of PF-ODE sub-paths divided by sampled points, and harnesses sub-path linear (SL) ODEs to form a progressive and continuous error estimation along each individual PF-ODE sub-path. The optimization on such SL-ODEs allows SPLAM to construct denoising mapping with smaller cumulative approximated error. An efficient distillation method is also developed to facilitate the incorporation of pre-trained diffusion models, such as latent diffusion models. The extensive experimental results demonstrate SPLAM achieves remarkable training efficiency, requiring only 6 A100 GPU days to produce a high-quality generative model capable of 2 to 4-step generation. Comprehensive evaluations on LAION, MS COCO 2014, and MS COCO 2017 datasets also illustrate that SPLAM surpasses the existing acceleration methods in few-step generation tasks, achieving state-of-the-art performance both on FID and the quality of the generated images.
  . 
\end{abstract}

\section{Introduction}
\label{sec:intro}
Diffusion models, also known as score-based generative models, have emerged as a potent paradigm in generative computer vision, enabling the synthesis of highly realistic images by progressively refining random noise into structured visual content~\cite{DDPM,score-base,Glide,dalle2,song2019generative}. Despite their impressive ability, one of the primary challenges associated with diffusion models lies in their computational intensity, often requiring hundreds of iteration steps to produce a single image. This has spurred a surge of research focused on accelerating diffusion models to retain high-quality outputs while significantly reducing the computation cost during the inference phase~\cite{DDIM, lu2022dpm, lu2022dpm++, progressivedistill, dmd, xu2023ufogen, liu2023instaflow, consistencymodels, LCM, liu2022flow}.

Within the spectrum of acceleration techniques, consistency models~\cite{consistencymodels,LCM} have garnered attention as they forge a consistent denoising mapping across points on Probability Flow (PF) ODE trajectories. The learning strategy brings consistency models a notable \textit{consistency property} and could estimate the overall prediction errors as a summation of incremental errors, which are computed as the difference between the predicted results of adjacent trajectory points. In this paper, we recognize that the approximation of denoising mappings by consistency models is essentially a minimization process targeting the endpoints of sub-paths along ODE trajectories. We observe that the approximated performance is currently limited by the accumulation of errors that arise from either an overabundance of approximation operations, or the heightened challenge of optimizing individual sub-path errors as the skipping step size expands. 

To address these challenges, we propose a novel approach in this paper, designated as the Sub-Path Linear Approximation Model (SPLAM). SPLAM adheres to the foundational concept of cumulative approximation of PF-ODE trajectories but innovates through its sustained learning from Sub-Path Linear (SL) ODEs. Specifically, we dissect the sub-path learning objective based on the noise prediction design~\cite{DDPM, EDM} into two interrelated aspects, and establish the SL-ODEs to give respective progressive or continuous estimation for each component, by a carefully designed linear interpolation between the endpoints of sub-paths. We then utilize the SL-ODEs to approximate the complete PF-ODE trajectories which allows a more nuanced optimization. Consequently, the prediction error of our approach is assessed through iterative solutions of all SL-ODEs, enabling a reduction of cumulative errors and an enhancement in image generation quality. Furthermore, we also develop an efficient distillation procedure for our SPLAM which enables the incorporation with pre-trained latent diffusion models \cite{LDM} (\eg, Stable Diffusion).  Our contributions can be summarized as below:
\begin{enumerate}
    \item We identify that the optimization process for consistency models essentially minimizes the cumulative approximated error along PF-ODE sub-path endpoints, and observe that the performance of such approximations is hindered by the proliferating number of approximations or the amplified difficulty in optimizing single sub-path errors for as skipping step size increases.
    \item To address these challenges, we propose a novel approach as Sub-Path Linear Approximation Model (SPLAM). SPLAM employs Sub-Path Linear (SL) ODEs to continuously approximate the complete PF-ODE trajectories and progressively optimize the sub-path learning objectives, which could construct the denoising mappings with smaller cumulative approximated errors. 
    \item Leveraging the proposed SPLAM and SL-ODE framework, we put forth an efficient distillation method. When integrated with powerful pre-trained models like Stable Diffusion, our approach allows more efficient training and respectively attains impressive FIDs as 10.09, 10.06, 20.77 in LAION, MS COCO 2014, MS COCO 2017 datasets, achieving better performance and close inference latency to all previous accelerating approaches. 
    \end{enumerate}
\begin{figure}[t]
  \centering
  \includegraphics[width=0.98\textwidth]{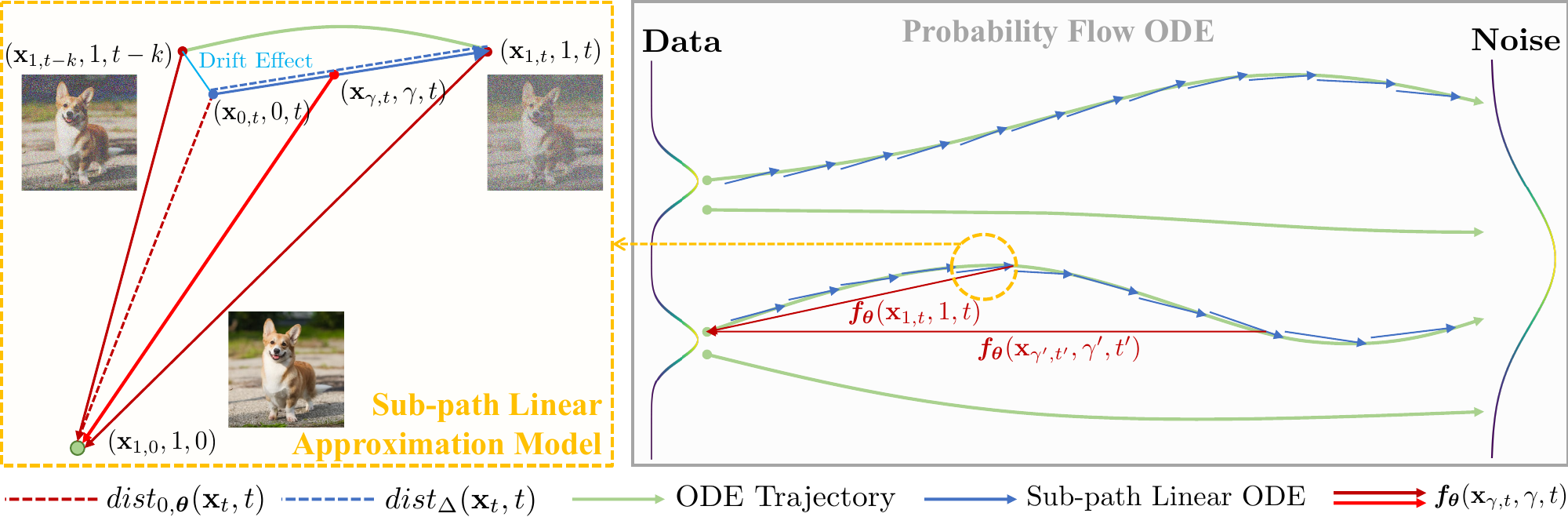}
  \caption{Our Sub-Path Linear Approximation Model employs Sub-Path Linear ODEs to approximate the sub-paths on the PF-ODE trajectories, which is determined by the linear interpolation of corresponding endpoints. SPLAM is then trained based on the consistent mapping along SL-ODEs to minimize the approximated errors.}
  \label{fig:method}
\end{figure}

\section{Related Work}
\noindent\textbf{Diffusion Models}~\cite{DDPM, balaji2022ediffi, score-base, sohl2015deep, EDM, improveddenosing, LDM} have solidified their status as a cornerstone in the realm of generative models, outshining previous approaches in creating rich and detailed images. Song \etal~\cite{score-base} model this process from continuous-time perspective with a stochastic differential equation (SDE), which iteratively denoise an initial noise distribution leveraging the learned \textit{score} of the data distribution to steer the process towards data points \cite{DDPM, song2019generative, score-base}.
This reverse diffusion process has been verified to be particularly adept at capturing the intricate structures and variations inherent in complex data sets.
They also demonstrate that there exists an ordinary differential equation (ODE), dubbed as Probability Flow (PF) ODE, which shares the marginal probability densities with the reverse-time SDE and thus yields a deterministic sampling trajectory \cite{EDM, score-base}.
In contrast to other generative models like VAEs \cite{kingma2013auto, sohn2015learning} and GANs \cite{Goodfellow2014GAN}, diffusion models demonstrate remarkable robustness in training and excel in producing samples with substantial diversity and high fidelity, thereby offering a robust solution for modeling complex distributions in an ever-expanding array of generative tasks.

\noindent\textbf{Accelerating Diffusion Models.}
While diffusion models have demonstrated their superiority in generating high-quality samples, the generation speed remains a major hindrance due to requiring thousands of sampling steps, which poses difficulties for practical and efficient applications.
To address these issues, a surge of advancements has emerged aiming to accelerate the inference process.
Some works concentrate on designing non-training fast diffusion samplers \cite{lu2022dpm, lu2022dpm++, score-base, EDM, liu2022pseudo, bao2022analytic, zhou2023fast, jolicoeur2021gotta}, potentially cutting down the steps from one thousand to a modest 20-50.
In the realm of distillation \cite{hinton2015distilling}, efforts have been undertaken \cite{progressivedistill, zhou2023fast, luhman2021knowledge, OnDistillation, berthelot2023tract, gu2023boot, zheng2022fast} to condense the inference steps of pre-trained diffusion models to fewer than 10.
Progressive distillation (PD) \cite{progressivedistill} seeks to amortize the integration of PF-ODE into a new sampler that takes half as many sampling steps, displaying efficacy with as few as 2/4 steps.
Consistency models \cite{consistencymodels, LCM, lcmlora, song2023improved}, as a nascent class of models, offer the promise of high-quality one-step generation by mapping any point along the PF-ODE trajectory back to the origin.
Representing flow-based approaches \cite{liu2023instaflow, liu2022flow, lipman2022flow, tong2023improving}, InstaFlow~\cite{liu2023instaflow, liu2022flow} propose a reflow technique to straighten the trajectories of probability flows and refine the coupling between noises and images, which achieves a one-step SD model.
Concurrently, some strategies are exploring the inclusion of GAN-like objectives into diffusion models to afford fast generative capabilities\cite{xu2023ufogen, sauer2023adversarial, sdxllightning, dmd}.
DMD \cite{dmd} additionally proposes a distribution matching method that enables one-step high-quality image generation.

\section{Preliminaries}

\noindent\textbf{Diffusion Models} are a class of generative models that gradually transform data into a noisy state through Gaussian perturbations and subsequently learn to reverse this process to reconstruct the original data by progressively denoising it. Denote $\bm{x}_0$ as the data sampled from the original distribution $\bm{x}_0\sim p_{data}(\bm{x})$ and $\alpha(t),\sigma(t)$ as functions that define a noise schedule. Diffusion models transition the data to a noise-corrupted marginal distribution, which can be expressed as:
\begin{equation}
    p_t(\bm{x}_t|\bm{x}_0)=\mathcal{N}(\bm{x}_t|\alpha(t)\bm{x}_0,\sigma(t)^2\mathbf{\mathit I}),
\end{equation}
for any time step $t\in [0,T]$.

Song \etal~\cite{score-base} describe the diffusion process using a stochastic differential equation (SDE):
\begin{equation}
    d\bm{x}_t=\bm{f}(\bm{x}_t,t)dt + g(t)d\bm{w}_t,
\end{equation}
where $f(\cdot,\cdot)$ and $g(\cdot)$ denote the drift and diffusion coefficients, respectively, and $\bm{w}_t$ signifies the standard Brownian motion at time $t$. They also derive an ordinary differential equation (ODE) corresponding to this SDE, which defines the trajectories of solutions sampled at time $t$ according to $p_t(\bm{x}_t)$:
\begin{equation}
    d\bm{x}_t=\left[\bm{f}(\bm{x}_t,t) - \frac{1}{2}g(t)^2\nabla_{\bm{x}} \log p_t(\bm{x}_t)\right] dt,
    \label{eq.ODEFunc}
\end{equation}
referred to as the Probability Flow (PF) ODE. In the reverse denoising process, models are taught to learn a score function $\mathbf{s}_{\bm{\theta}}(\bm{x}_t,t)\approx \nabla\log p_t(\bm{x}_t)$, adhering to the PF-ODE. Therefore, diffusion models are also recognized as score-based generative models. Based on the diffusion process, latent diffusion models (LDMs) additionally employ a VAE encoder $\mathcal{E}(\cdot)$ and decoder $\mathcal{D}(\cdot)$ to compress the image $\bm{x}$ into latent space as $\bm{z}=\mathcal{E}(\bm{x})$ and reconstruct it by the decoder: $\hat{\bm{x}}=\mathcal{D}(\bm{z})$, and implement the diffusion process on the compressed vector $\bm{z}$ via latent space~\cite{LDM}. With the latent diffusion process, the pre-trained large-scale LDMs like Stable Diffusion (SD) Models could achieve more precise PF-ODE solutions and thus generate high-quality images.

\noindent\textbf{Consistency Model} has been proposed by Song \etal\cite{consistencymodels} as a novel paradigm within the family of generative models. Considering a solution trajectory of the PF-ODE $\{(\bm{x}_t,t)\}_{t\in [\epsilon,T]}$, consistency models comply with a \textit{consistency function} that projects every pair $(\bm{x}_t,t)$ along the trajectory back to the starting point: $\bm{F}(\bm{x}_t,t)\mapsto \bm{x}_\epsilon$, for any $t\in [\epsilon,T]$, to obtain a one-step generator. Here, $\epsilon$ represents a small positive constant, thereby making $\bm{x}_\epsilon$ a viable surrogate for $\bm{x}_0$.
An important characteristic of the consistency models is the \textit{self-consistency property}:
\begin{equation}
    \bm{F}(\bm{x}_t,t)=\bm{F}(\bm{x}_t',t'),\quad \forall t, t' \in [\epsilon,T],
\end{equation}
which is leveraged as the training constraint for the consistency models, whether when distilling knowledge from a pre-trained model or training from scratch. 
The model is parameterized as follows:
\begin{equation}
    \bm{F}_{\bm{\theta}}(\bm{x}_t,t) = c_{\mathrm{skip}}(t)\bm{x}_t+c_{\mathrm{out}}(t)\bm{f}_{\bm{\theta}}(\bm{x}_t,t),
\end{equation}
where $c_{\text{skip}}(t)$ and $c_{\text{out}}(t)$ are differentiable functions ensuring that $c_{\text{skip}}(\epsilon)=1$ and $c_{\text{out}}(\epsilon)=0$, guaranteeing that $\bm{F}_{\bm{\theta}}(\bm{x}_\epsilon,\epsilon)\equiv \bm{x}_\epsilon$, and $\bm{f}_{\bm{\theta}}(\bm{x}_t,t)$ is a deep neural network. For the distillation approach called as \textit{Consistency Distillation}, the training objective is formulated as:
\begin{equation}
    \mathcal{L}_{CD}({\bm{\theta}},{\bm{\theta}}^-;\phi)=\mathbb{E}[d(\bm{F}_{\bm{\theta}}(\bm{x}_{t_{n+1}},t_{n+1}),\bm{F}_{{\bm{\theta}}^-}(\hat{\bm{x}}^\Phi_{t_{n}},t_n))],
    \label{eq.CDLoss}
\end{equation}
where $\hat{\bm{x}}^\Phi_{t_n}=\bm{x}_{t_{n+1}}+(t_{n+1}-t_n)\Phi(\bm{x}_{t_{n+1}},t_{n+1};\phi)$ serves as a one-step estimation of $\bm{x}_{t_n}$ based on $\bm{x}_{t_{n+1}}$ from $\Phi(\cdot;\phi)$, a update function of a one-step ODE solver, and $d(\cdot,\cdot)$ is a chosen distance metric. Consistency models also utilize the EMA strategy to stabilize the training, and ${\bm{\theta}}^-$ is the running average of ${\bm{\theta}}$. Latent Consistency Models (LCMs)~\cite{LCM} introduce consistency model into the distillation for latent diffusion models. To accelerate the training of consistency models, LCM employs a skipping step size $k$ to ensure consistency between the current timestep and $k$-step away. With a conditional input $c$ and a guidance scale $w$ to achieve the CFG strategy~\cite{CFG}, the modified learning objective for the \textit{latent consistency distillation} is formulated as:
\begin{equation}
    \mathcal{L}_{LCD}({\bm{\theta}},{\bm{\theta}}^-;\phi)=\mathbb{E}[d(\bm{F}_{\bm{\theta}}(\bm{x}_{t_{n+k}},w,c,t_{n+k}),\bm{F}_{{\bm{\theta}}^-}(\hat{\bm{x}}^\Phi_{t_{n}},w,c,t_n))].
\end{equation}
\section{Methodology}
\subsection{Approximation Strategy for Denoiser}

\noindent\textbf{One-step Denoiser Parameterization.} To synthesize an image from a sampled input $\bm{x}_t$ at a large time step $t$ in one-step, a natural approach is to adopt the strategy from \cite{DDPM} that employs a neural network $\bm{\epsilon}_{\bm{\theta}}(\bm{x}_t,t)$ to predict a standard Gaussian distribution, which implements the denoising mapping parameterized as $\bm{f}_{\bm{\theta}}(\bm{x}_t,t)=\frac{\bm{x}_t - \sigma(t)\bm{\epsilon}_{\bm{\theta}}(\bm{x}_t,t)}{\alpha(t)}$.  By redefining the target distribution for $(\bm{x}_t,t)$ as $\bm{x}^t_0=\alpha(t)\bm{x}_0\sim p_{data,t}(\alpha(t)\bm{x})$ and $\bm{D}_{\bm{\theta}}(\bm{x}_t,t)=\alpha(t)*\bm{f}_{\bm{\theta}}(\bm{x}_t,t)=\bm{x}_t-\sigma(t)\bm{\epsilon}_{\bm{\theta}}(\bm{x}_t,t)$, this predictive formulation can be recast into a canonical denoiser function defined in~\cite{EDM} that aims to minimize the denoising error as follows:
\begin{equation}
    \mathcal{L}_{\bm{D}}({\bm{\theta}})= \mathbb{E}_{\bm{x}^t_0\sim p_{data,t},\bm{x}_t\sim \mathcal{N}(\bm{x}^t_0,\sigma(t)^2I)}[|\bm{D}_{\bm{\theta}}(\bm{x}_t,t)-\alpha(t)\bm{x}_0|],
    \label{eq.DenoiserLoss}
\end{equation}
where $|\cdot|$ is an estimation of the error vector (\eg, a L2 distance). \textit{However, the \cref{eq.DenoiserLoss} is hard to be optimized in practice.} For instance, when $\alpha(t)$ decreases over time step $t$ which implies $\alpha(t)\bm{x}_0 \to \mathbf{0}$, the training is likely to collapse and the denoiser is taught to generally give a zero output.

\noindent\textbf{Approximation Strategy in Consistency Models.} 
We observe that, consistency models~\cite{consistencymodels,LCM} provide a solution to the aforementioned issues by leveraging the \textit{consistency property}. As we presume that we have obtained a good prediction result $\bm{f}_{\bm{\theta}}(\bm{x}_{t-k})\approx \bm{x}_0$, from a time step $t-k$ ahead of $t$ for $k$ steps, this property yields an approximated error estimation of \cref{eq.DenoiserLoss} as :
\begin{equation}
    \mathbb{E}[|\bm{D}_{\bm{\theta}}(\bm{x}_t,t)-\alpha(t)\bm{f}_{\bm{\theta}}(\bm{x}_{t-k},t-k)|].
    \label{eq.DenoiserApprox}
\end{equation}
By incorporating the expressions for $\bm{f}_{\bm{\theta}}(\bm{x}_{t-k},t-k)$ and $\bm{D}_{\bm{\theta}}(\bm{x}_t,t)$, we derive the approximated error estimation based on $\bm{\epsilon}_{\bm{\theta}}(\cdot,\cdot)$ as:
\begin{equation}
    \small
    \mathcal{L}_{\text{Approx}}({\bm{\theta}})= \mathbb{E}[|\bm{x}_t-\frac{\alpha(t)}{\alpha(t-k)}\bm{x}_{t-k}+\frac{\alpha(t)}{\alpha(t-k)}\sigma(t-k)\bm{\epsilon}_{\bm{\theta}}(\bm{x}_{t-k},t-k)-\sigma(t)\bm{\epsilon}_{\bm{\theta}}(\bm{x}_t,t)|],
    \label{eq.DenoiserCM}
\end{equation}
where the mentioned impact on optimization is reduced as the coefficient is amplified by $\alpha(t-k)$. When $k$ is limited to 1, the error between the mapping result $\bm{f}_{\bm{\theta}}(\bm{x}_t,t)$ and the trajectory origin $\bm{x}_0$ can be quantified by the accumulation of incremental approximated errors~\cite{consistencymodels}: $|\bm{x}_0-\bm{f}_{\bm{\theta}}(\bm{x}_t,t)|\leq \sum\limits_{ 1 \leq t'\leq t} |\bm{f}_{\bm{\theta}}(\bm{x}_{t'},t')-\bm{f}_{\bm{\theta}}(\bm{x}_{t'-1},t'-1)|$. Ideally if the error of one single approximation can be bounded, we can reduce the cumulative error by decreasing the number of approximations. This technique, also called \textsc{Skipping-Step} in LCM~\cite{LCM}, extends to optimize the error for skipping sampled points on the trajectories as $|\bm{f}_{\bm{\theta}}(\bm{x}_{t'},t')-\bm{f}_{\bm{\theta}}(\bm{x}_{t'-k},t'-k)|$ for a fixed skipping step size $k$. However, our insights reveal this precondition does not hold for extended situations. Denote $\{\bm{x}_{t'}\}_{t'\in[t-k,t]}$ as the \textit{sub-path} between $\bm{x}_{t-k}$ and $\bm{x}_t$ from the original PF-ODE trajectory, we discern that the learning objective in \cref{eq.DenoiserCM} for $\bm{\epsilon}_{\bm{\theta}}(\bm{x}_t,t)$ can be decomposed into two complementary components: 1) $dist_{\Delta}(\bm{x}_{t-k},\bm{x}_t,t)=\bm{x}_t-\frac{\alpha(t)}{\alpha(t-k)}\bm{x}_{t-k}$, which gauges the incremental distance from $\bm{x}_{t-k}$ to $\bm{x}_t$ attributable to the drift and diffusion processes, and 2) $dist_{0,{\bm{\theta}}}(\bm{x}_{t-k},t-k,t)=\frac{\alpha(t)}{\alpha(t-k)}\sigma(t-k)\bm{\epsilon}_{\bm{\theta}}(\bm{x}_{t-k},t-k)$, which captures the denoising contribution from previous time steps that should be coherently propagated to subsequent time steps $t$. Thus we rewrite \cref{eq.DenoiserCM} as a \textit{sub-path learning objective}:
\begin{equation}
    \begin{aligned}
        \mathcal{L}_{\text{Sub-p}}({\bm{\theta}},k)& =\mathbb{E}[|dist_\Delta(\bm{x}_t,\bm{x}_{t-k},t)+dist_{0,{\bm{\theta}}}(\bm{x}_{t-k},t-k,t)-\sigma(t)\bm{\epsilon}_{\bm{\theta}}(\bm{x}_t,t)|]. \\
    \end{aligned}
    \label{eq.DenoiserSubPath}
\end{equation}

In \cref{eq.DenoiserSubPath}, the learning process of $dist_\Delta$ equates to modeling the denoising distribution $p(\bm{x}_{t-k}|\bm{x}_t)$, which deviates from Gaussian for larger skipping step sizes and is found to be intractable to estimate~\cite{SIDDM,lu2022dpm,lu2022dpm++,xu2023ufogen,EDM}. Consequently, the approximated error escalates uncontrollably with increased $k$ due to reliance on the flawed learning. Although LCM sets an empirical $k$ of 20 to balance the pros and cons, the fundamental issues remain unaddressed and unexplored.

\subsection{Sub-Path Linear Approximation Model}

To improve the learning objective in \cref{eq.DenoiserSubPath}, in this paper we introduce a new approach for accelerating diffusion models termed \textbf{S}ub-\textbf{p}ath \textbf{L}inear \textbf{A}pproximation \textbf{M}odel (SPLAM).  SPLAM introduces Sub-Path Linear (SL) ODEs to approximate the sub-paths on the PF-ODE trajectories as a linear interpolation between the according sub-path endpoints. As the optimization based on such SL-ODEs gives a respectively progressive and continuous estimation for the decomposed two terms in \cref{eq.DenoiserSubPath}, our SPLAM is trained based on the conducted SL-ODE learning objectives, and achieves smaller overall prediction errors and better generation quality. We also develop an efficient distillation procedure for latent diffusion models
~\cite{LDM}, with \textit{Multiple Estimation} strategy which improves the estimated results of teacher models. 

\subsubsection{Sub-Path Linear ODE}
\label{sec.SAODE}
Based on the above analysis, in this paper, we introduce \textit{Sub-Path Linear} (SL) ODEs to model approximated sub-paths in the original PF-ODE trajectories, which gives a progressive estimation for $dist_\Delta$. For a sampled sub-path $\{\bm{x}_{t'}\}_{t\in[t-k,t]}$ on a solution trajectory dictated by \cref{eq.ODEFunc}, we interpolate a linear path from $(\bm{x}_{t-k},t-k)$ to $(\bm{x}_t,t)$, guided by the vector direction of $dist_\Delta(\bm{x}_t,\bm{x}_{t-k},t)$. To distinguish the impacts of $dist_\Delta$ and $dist_{0,{\bm{\theta}}}$, we account for the drift component in the linear approximated path, causing a shift on coefficient from $(\bm{x}_{t-k},t-k)$ to $(\frac{\alpha(t)}{\alpha(t-k)}\bm{x}_{t-k},t-k)$. The points on the approximated path $\{\bm{x}_{\gamma,t}\}_{\gamma\in[0,1]}$ are thus computed as:
\begin{equation}
    \begin{aligned}
        \bm{x}_{\gamma,t}&=\frac{\alpha(t)}{\alpha(t-k)}\bm{x}_{t-k}+\gamma * dist_\Delta(\bm{x}_{t},\bm{x}_{t-k},t)\\
                      &=(1-\gamma)\frac{\alpha(t)}{\alpha(t-k)}\bm{x}_{t-k}+\gamma \bm{x}_{t},
    \end{aligned}
    \label{eq.SAPath}
\end{equation}
for a sampled $(\bm{x}_{t-k},t-k)$ and $(\bm{x}_t,t)$.

Since $\bm{x}_t$ and $\bm{x}_{t-k}$ conform to distributions governed by the PF-ODE, our linear transformation effectively defines a linear ODE from distribution $\frac{\alpha(t)}{\alpha(t-k)}\bm{x}_{t-k}\sim p_{t-k,k}(\bm{x}_{t-k})$ to $\bm{x}_{t}\sim p_{t}(\bm{x}_{t})$ over $\gamma$, where $p_{t,k}(\bm{x}_t)$ has the property $p_{t,k}(\bm{x}_t|\bm{x}_0)=\mathcal{N}(\alpha(t+k)\bm{x}_0,\left[\frac{\alpha(t+k)\sigma(t)}{\alpha(t)}\right]^2 I)$:
\begin{equation}
    d\bm{x}_{\gamma,t}=[\gamma*dist_\Delta(\bm{x}_{t},\bm{x}_{t-k},t)]d\gamma.
\end{equation}
We denote it as \textit{Sub-Path Linear} (SL) ODE. To apply the approximation strategy on the SL-ODE, the Denoiser and generation function replacing $\bm{x}_t$ with $\bm{x}_{\gamma,t}$ are given by:
\begin{equation}
    \begin{aligned}
    &\bm{D}_{\bm{\theta}}(\bm{x}_{\gamma,t},\gamma,t) = \bm{x}_{\gamma,t}-\sigma(\gamma,t)\bm{\epsilon}_{\bm{\theta}}(\bm{x}_{\gamma,t},\gamma,t),\\
    &\bm{f}_{\bm{\theta}}(\bm{x}_{\gamma,t},\gamma,t) = \frac{\bm{D}_{\bm{\theta}}(\bm{x}_{\gamma,t},\gamma,t)}{\alpha(t)}.
    \end{aligned}
    \label{eq.FuncSAODE}
    \end{equation}
Incorporating these into \cref{eq.DenoiserSubPath}, we derive the sub-path learning objective for our SPLAM model as :
\begin{equation}
\small
\mathcal{L}_{\text{SPLAM}}({\bm{\theta}},k)=\mathbb{E}[|\gamma * dist_\Delta(\bm{x}_{t},\bm{x}_{t-k},t)+dist_{0,{\bm{\theta}}}(\bm{x}_{t-k},t,t-k)-\sigma(\gamma, t)\bm{\epsilon}_{\bm{\theta}}(\bm{x}_{\gamma,t}, \gamma, t)|],
\label{eq.DenoiserSPLAM}
\end{equation}
which gives a progressive estimation for the otherwise intractable $dist_\Delta$ objective. The value for $\sigma(\gamma,t)$ can be precisely estimated from the distribution $p_t(\bm{x}_t)$ and $p_{t-k}(\bm{x}_{t-k})$ but has a complex expression. Empirically we utilize an approximate result as $\sigma(\gamma,t) = (1-\gamma)\frac{\alpha(t)}{\alpha(t-k)}\sigma(t-k)+\gamma * \sigma(t)$. Compared to consistency models which adopt \cref{eq.DenoiserCM} or \cref{eq.DenoiserSubPath}, our $\mathcal{L}_{\text{}}$ maintains a progressive estimation for $dist_\Delta$ and a consistent estimation for $dist_{0,{\bm{\theta}}}$, which enables the learning for large skipping step size. The overall prediction error can still be assessed by the aggregate of approximated errors between the sub-path endpoints and the approximated error between these points is continuously optimized through the SL-ODEs. Consequently, the optimization for the approximated errors in our SPLAM could be significantly improved. Our approach could further benefit from the increased skipping step size, allowing our method to generate images of higher quality with reduced sampling steps in more efficient training. 

\subsubsection{Sub-Path Linear Approximation Distillation}
\label{subsec:sld}
 In this paper, we adopt pre-trained Stable Diffusion (SD) models~\cite{LDM} to obtain the solution PF-ODE trajectories which we build our SL-ODEs upon, and we call the approach \textbf{S}ub-\textbf{p}ath \textbf{L}inear \textbf{A}pproximation \textbf{D}istillation (SPLAD). To achieve conditional generation with the conditional input $c$, the noise prediction model is parameterized as $\bm{\epsilon}_{\bm{\theta}}(\bm{z}_t,c,t)$~\cite{score-base,lu2022dpm}. We also introduce $\gamma$ into the prediction models for solving our SL-ODEs, and leverage the $\gamma-$conditioned training where $\gamma$ is converted to Fourier embeddings and fed into the models as an input. Specifically, to predict $\bm{z}_0$ in the latent space, the generation function for SPLAM is defined as:
\begin{equation}
\bm{F}_{\bm{\theta}}(\bm{z}_{\gamma,t},c,\gamma,t)=c_{\text{skip}}(t)\bm{z}_{\gamma,t}+c_{\text{out}}(t)\bm{f}_{\bm{\theta}}(\bm{z}_{\gamma,t},c,\gamma,t),
\end{equation}
where $\bm{f}_{\bm{\theta}}(\bm{z}_{\gamma,t},c,\gamma,t)$ mirrors \cref{eq.FuncSAODE} while replacing $\bm{\epsilon}_{\bm{\theta}}(\bm{z}_{\gamma,t},\gamma,t)$ with the conditional form $\bm{\epsilon}_{\bm{\theta}}(\bm{z}_{\gamma,t},c,\gamma,t)$. The functions $c_{\text{skip}}$ and $c_{\text{out}}$ are leveraged to ensure that $\bm{F}_{\bm{\theta}}(\bm{z}_{1,0},c,1,0)\equiv \bm{z}_0$ (we regard $F_{\bm{\theta}}$ as the same expression of $f_{\bm{\theta}}$ since $c_{\text{skip}}(t)\ll c_{\text{out}}(t)$ for most time steps). Integrating this with \cref{eq.DenoiserApprox}, our SPLAD approach minimizes the following objective:
\begin{equation}
    \small
    \mathcal{L}_{\text{SPLAD}}({\bm{\theta}},{\bm{\theta}}^-;\phi)=\mathbb{E}_{\bm{z}_0\sim p_{data},t\sim\mathcal{U}[k,T],\gamma\sim\mathcal{U}[0,1]}[|\bm{F}_{\bm{\theta}}(\bm{z}_{\gamma,t},c,\gamma,t)-\bm{F}_{{\bm{\theta}}^-}(\hat{\bm{z}}^\Phi_{1,t-k},c,1,t-k)|],
    \label{eq.LossSPLAD}
\end{equation}
where $\mathcal{U}$ denotes the uniform distribution, and $k$ is a pre-determined skipping step size. The $\alpha(t)$ in \cref{eq.DenoiserApprox} is omitted due to its negligible effect on optimization in practice. The term $\hat{\bm{z}}^\Phi_{1,t-k}=\hat{\bm{z}}^\Phi_{t-k}$ is estimated using ODE solvers $\Phi(\cdot\cdot\cdot;\phi)$ derived from teacher models. In this paper DDIM~\cite{DDIM} is employed as our choice from the advanced solvers of LDMs. Moreover, to improve the estimation of $\hat{\bm{z}}^\Phi_{t-k}$, we apply the \textit{Multiple Estimation} which executes the solver $\Phi(\cdot\cdot\cdot,\phi)$ multiple times with a reduced skipping step size $k_\phi$. Denoting $t_{\phi,i}=t-i*k_\phi$ and initializing $\hat{\bm{z}}^\Phi_{t_{\phi,0}}=\bm{z}_t$, the multiple estimation is iteratively executed as:
\begin{equation}
    \hat{\bm{z}}^\Phi_{t_{\phi,i+1}}=\hat{\bm{z}}^\Phi_{t_{\phi,i}}+w\Phi(\hat{\bm{z}}^\Phi_{t_{\phi,i}},t_{\phi,i},t_{\phi,i+1},c;\phi)+(1-w)\Phi(\hat{\bm{z}}^\Phi_{t_{\phi,i}},t_{\phi,i},t_{\phi,i+1},\emptyset;\phi),
\end{equation}
for $i=0,1,2,...;i\leq \frac{k}{k_\phi}-1$, where $\emptyset$ denotes no conditional inputs and $w$ is a fixed guidance scale which controls the effect of conditional generation~\cite{CFG} from the conditional input $c$. 
\begin{algorithm}[!t]
    \small
    \caption{Sub-Path Linear Approximation Distillation (SPLAD)}
    \label{algo.SPLAD}
 \begin{algorithmic}
    \State {\bfseries Input:} dataset $\mathcal{D}$, initial model parameter ${\bm{\theta}}$, learning rate $\eta$, EMA decay rate $\mu$, ODE solver $\Phi(\cdot, \cdot; \phi)$, distance estimation $|\cdot|$, a fixed guidance scale $w$, step size $k$, VAE encoder $\mathcal{E}(\cdot)$, noise schedule $\alpha(t), \sigma(t)$
    \State ${\bm{\theta}}^-\leftarrow {\bm{\theta}}$ 
    \Repeat
    \State sample $(x,c) \sim \mathcal{D}, t \sim \mathcal{U}[k,T]$ and $\gamma\sim\mathcal{U}[0,1]$
    \State convert $x$ into latent space: $z=\mathcal{E}(x)$ 
    \State sample $\bm{z}_t\sim \mathcal{N}(\alpha(t)z,\sigma(z)^2I)$
    \State $\hat{\bm{z}}^\Phi_{t_{\phi,0}}\leftarrow \bm{z}_t$ , $i\leftarrow 0$
    \Repeat
    \State    $\hat{\bm{z}}^\Phi_{t_{\phi,i+1}}
\leftarrow\hat{\bm{z}}^\Phi_{t_{\phi,i}}+w\Phi(\hat{\bm{z}}^\Phi_{t_{\phi,i}},t_{\phi,i},t_{\phi,i+1},c;\phi)+(1-w)\Phi(\hat{\bm{z}}^\Phi_{t_{\phi,i}},t_{\phi,i},t_{\phi,i+1},\emptyset;\phi)$
    \State $i\leftarrow i+1$
    \Until{$k=i*k_\phi$}
    \State $\bm{z}_{\gamma,t}\leftarrow (1-\gamma) * \frac{\alpha(t)}{\alpha(t-k)}\hat{\bm{z}}^\Phi_{i-k}+\gamma * \bm{z}_t$
    \Comment{Sample a point on the SL-ODE.}
    \vspace{0.33em}
    \State $\mathcal{L}({\bm{\theta}},{\bm{\theta}}^-;\phi)\leftarrow  |(\bm{F}_{\bm{\theta}}(\bm{z}_{\gamma,t},c,\gamma,t)-\bm{F}_{\bm{\theta}}(\hat{\bm{z}}^\Phi_{1,t-k},c,1,t-k))|$
    \State ${\bm{\theta}}\leftarrow{\bm{\theta}}-\eta\nabla_{\bm{\theta}}\mathcal{L}({\bm{\theta}},{\bm{\theta}}^-;\phi)$
    \State ${\bm{\theta}}^-\leftarrow$ stopgrad($\mu{\bm{\theta}}^-+(1-\mu){\bm{\theta}}$)
    \vspace{0.33em}
    
    \Until{convergence}
 \end{algorithmic}
 \end{algorithm}
The pseudo-code for SPLAD is presented in \cref{algo.SPLAD}. SPLAD shares a similar training pipeline with consistency models\cite{consistencymodels,LCM} but can be distinguished as it optimizes the sub-path learning objectives based on the SL-ODEs and utilizes the $\gamma$-conditioned training. For a pair of input noise and time step $(\bm{z}_t,t)$, SPLAM gives the prediction of the denoised latent $\hat{\bm{z}}_0$ as:
\begin{equation}
    \hat{\bm{z}}_0 = \bm{F}_{{\bm{\theta}}^-}(\bm{z}_{1,t},c,1,t),
\end{equation}
for one-step generation, adhering strictly to the $\gamma=1$ condition. We also use the same iterative sample strategy as illustrated in \cite{consistencymodels} which could improve the quality of the generated images. In practice, we set the $\gamma\text{-}$embedding to $\mathbf{0}$ for $\gamma=1$, thereby allowing the weights associated with trained $\gamma$-embeddings to be discarded post-training. Thus our Sub-Path Linear Approximation Model (SPLAM) necessitates no additional parameters beyond the training phase and can be utilized the same as teacher models.

\section{Experiments}

In this section, we conduct experiments to examine the performance of our proposed Sub-Path Linear Approximation Model (SPLAM).
Firstly, we describe the experiment configuration and implementation details, and evaluate our models comprehensively on the text-to-image task (\cref{sec:exp1}).
Secondly, we verify the effectiveness of our algorithm design through detailed ablation studies (\cref{sec:exp2}).
Finally, we present the qualitative results of our SPLAM. (\cref{sec:exp3}).

\subsection{Text-to-Image Generation}
\label{sec:exp1}

\subsubsection{Experimental Configuration}
On text-to-image generation task, we train two models with pre-trained Stable Diffusion-V1.5 (SDv1.5) and Stable Diffusion-V2.1-base (SDv2.1-base) as teacher models respectively.
Following the setting of \cite{LCM}, the training dataset is one subset of LAION-5B\cite{schuhmann2022laion}: LAION-Aesthetics-6+.
We choose DDIM-Solver as the ODE solver $\phi$ at skipping step $k_\phi = 20$.

For evaluation, we adopt the commonly used FID and CLIP Score metrics.
The results are reported on both SDv1.5 and SDv2.1-base backbones, thus verifying the generalizability of our method.
For the experiment of distilling SDv2.1-base, we bench-mark our model on two test sets, LAION-Aesthetics-6+ as used in LCM \cite{LCM} and MSCOCO2014-30k for zero-shot generalization.
We also reproduce a SDv2.1-base LCM according to the training configuration outlined in \cite{LCM} while replacing the $w$-condition with the fixed guidance scale, which has also improved the performance.
We generally set the guidance scale for distilling SDv2.1-base to 8 and skipping step size to 20, which is consistent with ~\cite{LCM}.
For the experiment of distilling SDv1.5, we compare our model with state-of-the-art generative models including foundation diffusion models, GANs, and accelerated diffusion models. 
The guidance scale is set to $3$ to obtain the optimal FID, and we adopt the huber~\cite{song2023improved} loss for our SPLAD metric.
The skipping step size is set to 100 for SPLAM which has shown fast convergence.
We examine our method on two commonly used benchmarks, MSCOCO2014-30k and MSCOCO2017-5k. More implementation details are provided in the supplementary materials.

\setlength{\tabcolsep}{3pt}
\begin{table}[t]
    \centering
    \caption{Quantitative results for SDv2.1-base with $w=8$. The results of DDIM, DPM, DPM++ and LCM$^*$ for LAION test-set are derived from \cite{LCM}. LCM (fix $w$) is our reproduction conducted as stated in the paper. The results on COCO-30k are evaluated by us.}
    \label{tab:SD2.1Results}
    \resizebox{0.98\textwidth}{!}{
    \begin{tabular}{l|cccccc|cccccc}
    \toprule
    \multirow{2}{*}{Methods} & \multicolumn{6}{c|}{LAION-Aesthetics-6+} & \multicolumn{6}{c}{COCO-30k} 
    \\
    \cmidrule(lr){2-7}
    \cmidrule(lr){8-13}
    & \multicolumn{3}{c}{FID$(\downarrow)$} & \multicolumn{3}{c|}{CLIP-Score$(\uparrow)$} & \multicolumn{3}{c}{FID$(\downarrow)$} & \multicolumn{3}{c}{CLIP-Score$(\uparrow)$} \\
    \midrule
    
    & 1 Step  & 2 Steps  & 4 Steps & 1 Step & 2 Steps & 4 Steps  & 1 Step  & 2 Steps  & 4 Steps  & 1 Steps & 2 Steps & 4 Steps \\ 
    \midrule
    DDIM \cite{DDIM} & 183.29 & 81.05 & 22.38 & 6.03 & 14.13 & 25.89 & 431.26 & 229.44 & 32.77 & 2.88 & 7.72 & 28.76 \\
    DPM Solver \cite{lu2022dpm} & 185.78 & 72.81 & 18.53 & 6.35 & 15.10 & 26.64 & 206.37 & 73.87 & 22.04 & 10.56 & 22.87 & 31.18 \\
    DPM Solver++ \cite{lu2022dpm++} & 185.78 & 72.81 & 18.43 & 6.35 & 15.10 & 26.64 & 206.35 & 73.82 & 22.11 & 10.57 & 22.87 & 31.16 \\
    LCM$^*$ \cite{LCM}   & 35.36 & 13.31 & 11.10 & 24.14 & 27.83& 28.69 & - & - & - & - & - & - \\
    LCM (fix $w$) \cite{LCM} & 32.41 & 12.17 & 10.43 & 26.99 & 30.13 & 30.76 & 43.87 & 15.71 & 14.88 & 27.66 & 31.07 & 31.52 \\
    \midrule
    SPLAM & 32.64 & \textbf{12.06} & \textbf{10.09} & \textbf{27.13} & \textbf{30.18} & \textbf{30.76} & \textbf{40.52} & \textbf{14.59} & \textbf{13.81} & \textbf{27.83} & 31.00 & 31.45 \\
    \bottomrule
    \end{tabular}
    }
\end{table}

\subsubsection{Main Results}
\begin{table}[t]
    \caption{Quantitative results for SDv1.5. Baseline numbers are cited from \cite{dmd} and \cite{xu2023ufogen}. All the results of LCM are our reproduction whose performance is aligned as stated in the paper. $^\dag$ Results are evaluated by us using the released models.}
    \vspace{-0.5cm}
    \setlength{\tabcolsep}{6pt}
    \begin{subtable}[t]{0.52\textwidth}
        \centering
        \caption{Results on MSCOCO2014-30k, $w=3$.}
        \resizebox{0.98\textwidth}{!}{
            \begin{tabular}{llcc}
                \toprule
                Family & Methods & Latency$(\downarrow)$ & FID$(\downarrow)$ \\
                \midrule
                \multirow{11}{*}{Unaccelerated} & DALL-E \cite{dalle} & - & 27.5 \\
                                               & DALL-E2 \cite{dalle2} & - & 10.39 \\
                                               & Parti-750M \cite{parti} & - & 10.71 \\
                                               & Parti-3B \cite{parti} & 6.4s & 8.10 \\
                                               & Parti-20B \cite{parti} & - & 7.23 \\
                                               & Make-A-Scene \cite{make-a-scene} & 25.0s & 11.84 \\
                                               & Muse-3B \cite{chang2023muse} & 1.3 & 7.88 \\
                                               & GLIDE \cite{Glide} & 15.0s & 12.24 \\
                                               & LDM \cite{LDM} & 3.7s & 12.63 \\
                                               & Imagen \cite{imagen} & 9.1s & 7.27 \\
                                               & eDiff-I \cite{balaji2022ediffi} & 32.0s & 6.95 \\
                \midrule
                \multirow{3}{*}{GANs} & LAFITE \cite{lafite} & 0.02s & 26.94 \\
                                      & StyleGAN-T \cite{sauer2022stylegan} & 0.10s & 13.90 \\
                                      & GigaGAN \cite{gigagan} & 0.13s & 9.09 \\
                \midrule
                \multirow{11}{*}{\makecell[l]{Accelerated\\Diffusion}} & DPM++ (4step) \cite{lu2022dpm++} & 0.26s & 22.36 \\
                                                       & UniPC (4step) \cite{zhao2023unipc} & 0.26s & 19.57 \\
                                                       & LCM-LoRA (4step) \cite{lcmlora} & 0.19s & 23.62 \\
                                                       & InstaFlow-0.9B \cite{liu2023instaflow} & 0.09s & 13.10 \\
                                                       & InstaFlow-1.7B \cite{liu2023instaflow} & 0.12s & 11.83 \\
                                                       & UFOGen \cite{xu2023ufogen} & 0.09s & 12.78 \\
                                                       & DMD \cite{dmd} & 0.09s & 11.49 \\
                                                       & LCM (2step) \cite{LCM} & 0.12s & 14.29 \\
                                                       & \textbf{SPLAM} (2step) & 0.12s & 12.31 \\
                                                       & LCM (4step) \cite{LCM} & 0.19s & 10.68 \\
                                                       & \textbf{SPLAM} (4step) & 0.19s & \textbf{10.06} \\
                \midrule
                Teacher & SDv1.5 \cite{LDM}$^\dag$ & 2.59s & 8.03 \\
                \bottomrule
            \end{tabular}
            \label{subtab:sd1.5-30k-w3}
        }

    \end{subtable}
    \hfill
    \setlength{\tabcolsep}{8pt}
    \begin{subtable}[t]{0.48\textwidth}
        \centering
        \caption{Results on MSCOCO2017-5k, $w=3$.}
        \resizebox{0.98\textwidth}{!}{
            \begin{tabular}{lccc}
                \toprule
                Methods & \#Step & Latency$(\downarrow)$ & FID$(\downarrow)$ \\
                \midrule
                \multirow{2}{*}{DPM Solver++ \cite{lu2022dpm++}$^\dag$} & 4 & 0.21s & 35.0 \\
                                              & 8 & 0.34s & 21.0 \\
                \midrule
                \multirow{3}{*}{Progressive Distillation \cite{progressivedistill}} & 1 & 0.09s & 37.2 \\
                                                          & 2 & 0.13s & 26.0 \\
                                                          & 4 & 0.21s & 26.4 \\
                \midrule
                CFG-Aware Distillation \cite{snapfusion} & 8 & 0.34s & 24.2 \\
                \midrule
                InstaFlow-0.9B \cite{liu2023instaflow} & 1 & 0.09s & 23.4 \\
                InstaFlow-1.7B \cite{liu2023instaflow} & 1 & 0.12s & 22.4 \\
                \midrule
                UFOGen \cite{xu2023ufogen} & 1 & 0.09s & 22.5 \\
                \midrule
                \multirow{2}{*}{LCM \cite{LCM}} & 2 & 0.12s & 25.22 \\
                                            & 4 & 0.19s & 21.41 \\
                \midrule
                \multirow{2}{*}{\textbf{SPLAM}} & 2 & 0.12s & 23.07 \\
                                              & 4 & 0.19s & \textbf{20.77} \\
                \bottomrule
            \end{tabular}
            \label{subtab:sd1.5-5k-w3}
        }
                
        \centering
        \caption{Results on MSCOCO2014-30k, $w=8$.}
        \vspace{-0.11cm}
        \resizebox{0.98\textwidth}{!}{
            \begin{tabular}{llcc}
                \toprule
                Family & Methods & Latency$(\downarrow)$ & FID$(\downarrow)$ \\
                \midrule
                \multirow{8}{*}{\makecell[l]{Accelerated\\Diffusion}} & DPM++ (4step) & 0.26s & 22.44 \\
                                                       & UniPC (4step) \cite{zhao2023unipc} & 0.26s & 23.30 \\
                                                       & LCM-LoRA (4step) \cite{lcmlora} & 0.19s & 23.62 \\
                                                       & DMD \cite{dmd} & 0.09s & 14.93 \\
                                                       & LCM (2step) \cite{LCM}  \cite{LCM} & 0.12s & 15.56 \\
                                                       & \textbf{SPLAM} (2step) & 0.12s & 14.50 \\
                                                       & LCM (4step) \cite{LCM} \cite{LCM} & 0.19s & 14.53 \\
                                                       & \textbf{SPLAM} (4step) & 0.19s & \textbf{13.39} \\
                \midrule
                Teacher & SDv1.5 \cite{LDM}$^\dag$ & 2.59s & 13.05 \\
                \bottomrule
            \end{tabular}
            \label{subtab:sd1.5-30k-w8}
        }
    \end{subtable}
\label{tab:sd1.5results}
\end{table}
The results for SDv2.1-base are presented in \cref{tab:SD2.1Results}, we use DDIM \cite{DDIM}, DPM \cite{lu2022dpm}, DPM++ \cite{lu2022dpm++} and LCM \cite{LCM} as baselines.
It reveals that our SPLAM surpasses baseline methods nearly across both test sets, at each step, and on both FID and CLIP Score metrics.
We suppose that the close results on LAION are caused by overfitting, since the test set and train set are sourced from the same data collection.
For SDv1.5 under the guidance scale $w=3$, the quantitative results are demonstrated in \cref{subtab:sd1.5-30k-w3} and \cref{subtab:sd1.5-5k-w3}.
Our model with 4 steps gets FID-30k of 10.06 and FID-5k of 20.77, which outperforms all other accelerated diffusion models, including flow-based method InstaFlow \cite{liu2023instaflow} and techniques that introduce GAN objectives such as UFOGen \cite{xu2023ufogen} and DMD \cite{dmd}.
Furthermore, SPLAM showcases commensurate results with state-of-the-art foundation generative models such as DALL-E2 \cite{dalle2}.
Even in two steps, SPLAM has achieved a competitive performance of FID-30k 12.31 with parallel algorithms.
In practical scenarios, a higher guidance scale $w$ is typically favored to enhance the resultant image quality. Accordingly, we trained our SPLAM with $w$ set to 8 and bench-mark it against a range of advanced diffusion methodologies, as delineated in \cref{subtab:sd1.5-30k-w8}.
In this regime, SPLAM also demonstrates significant advantages, achieving state-of-the-art performance with a four-step FID-30k of 13.39 which exceeds other models by a large margin and is close to the teacher model.
Notably, the FID-30k of our model with only two steps reaches 14.50, surpassing the four-step LCM and DMD.
While DMD training consumes over one hundred A100 GPU days, which is more than 16 times our training duration.

\subsection{Ablation Study}
\label{sec:exp2}

\begin{figure}[t]
  \centering
  
  \begin{subfigure}{0.32\textwidth}
    \centering
    \includegraphics[width=\textwidth]{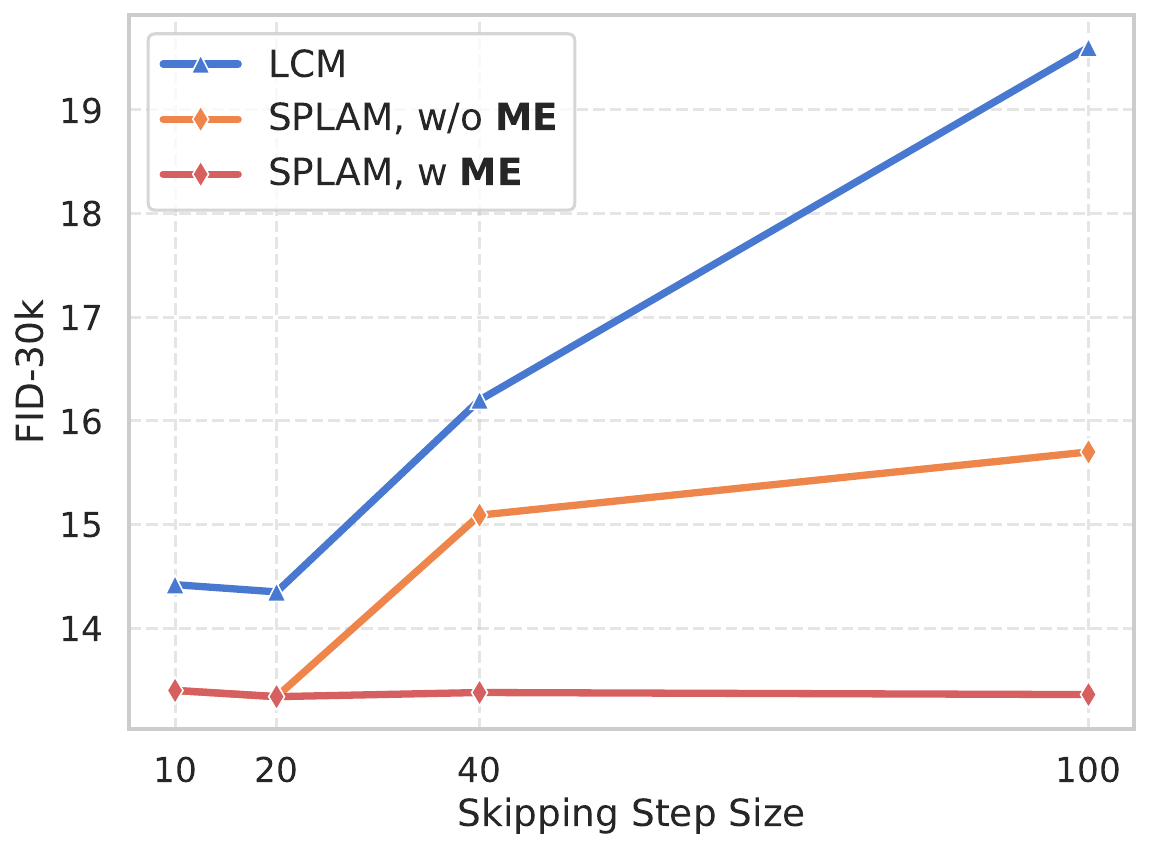}
    \caption{}
    \label{subfig:stepsize}
  \end{subfigure}
  \hfill
  \begin{subfigure}{0.32\textwidth}
    \centering
    \includegraphics[width=\textwidth]{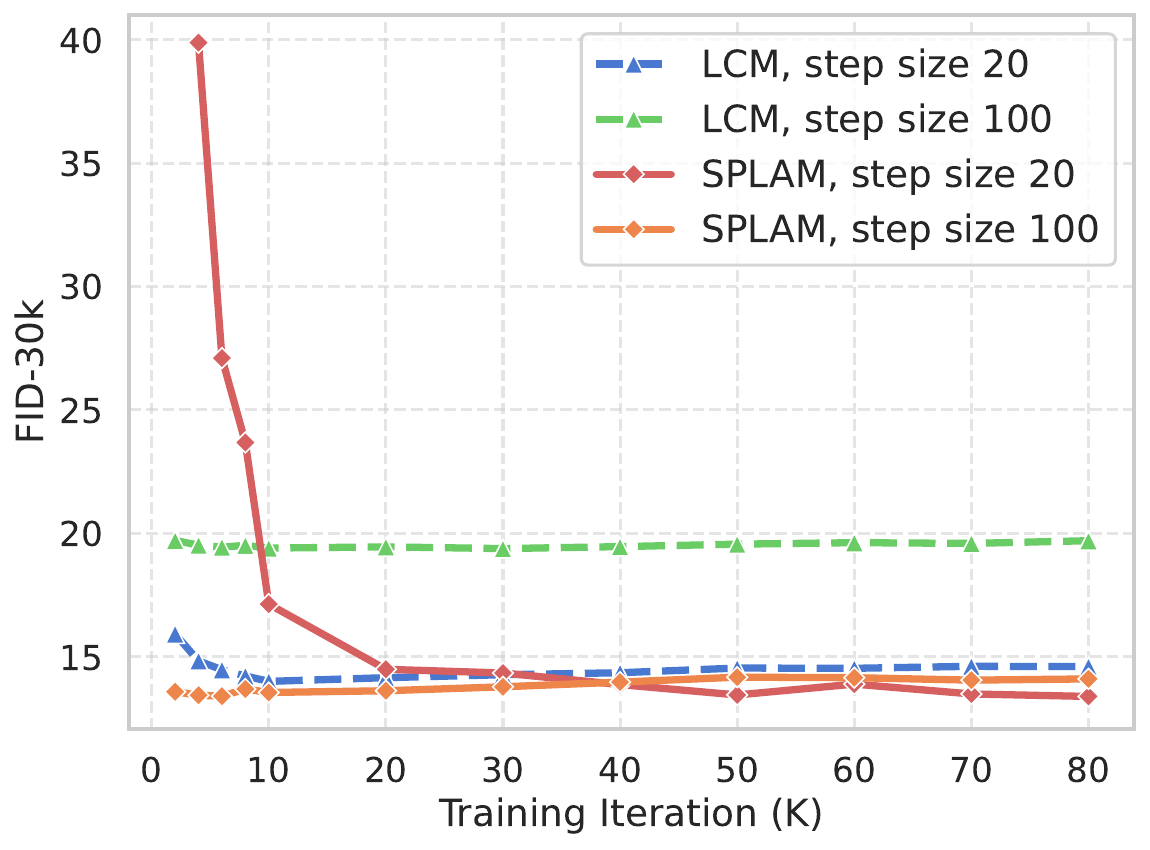}
    \caption{}
    \label{subfig:iteration}
  \end{subfigure}
  \hfill
  \begin{subfigure}{0.32\textwidth}
      \centering
      \includegraphics[width=\textwidth]{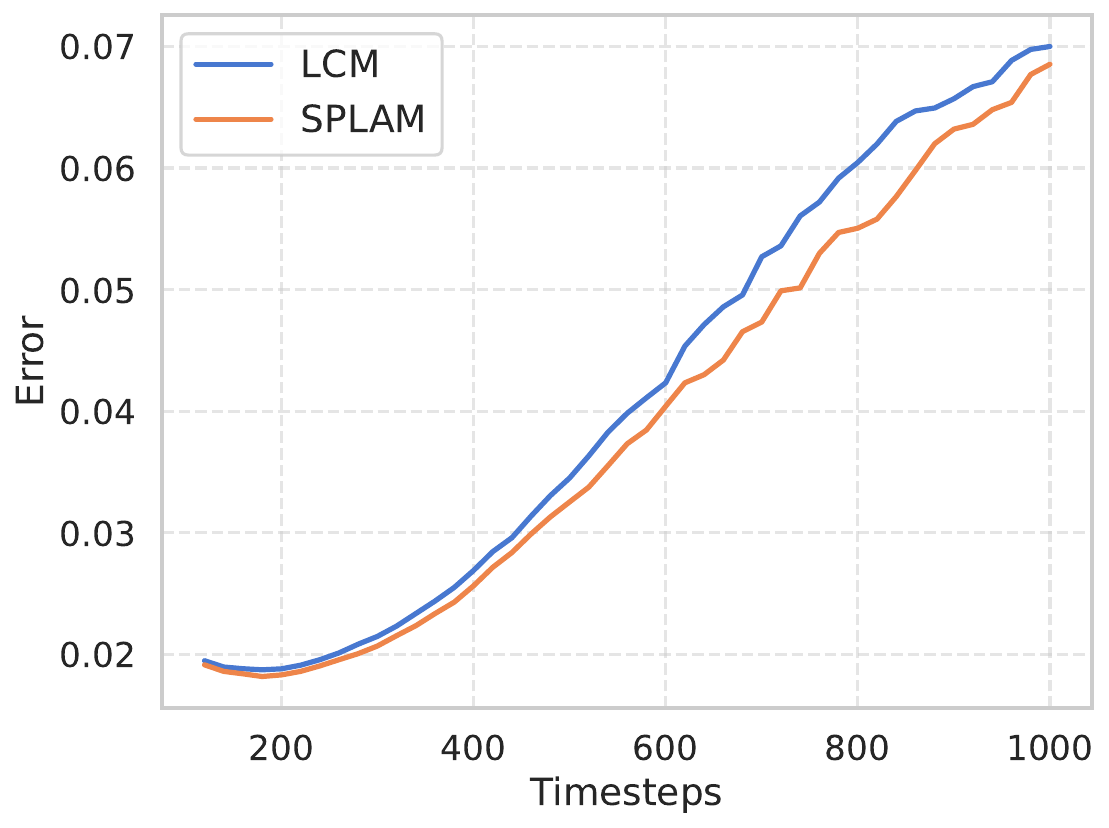}
      \caption{}
      \label{subfig:diff}
  \end{subfigure}
  \caption{(a) Ablations on skipping step size and skipping mechanism. \textbf{ME} denotes for our Multiple Estimation strategy. (b) Training curve comparing LCM and SPLAM. Our SPLAM with step size 100 is conducted with \textbf{ME}, which brings faster convergence. (c) Estimation of the error $\delta$ between consistency mapping values of two adjacent points through PF-ODE. SPLAM consistently outperforms LCM in terms of the error.}
  \label{fig:white}
\end{figure}
\begin{figure}[t]
  \centering

  \begin{subfigure}{.62\textwidth}
    \centering
    \includegraphics[width=\linewidth]{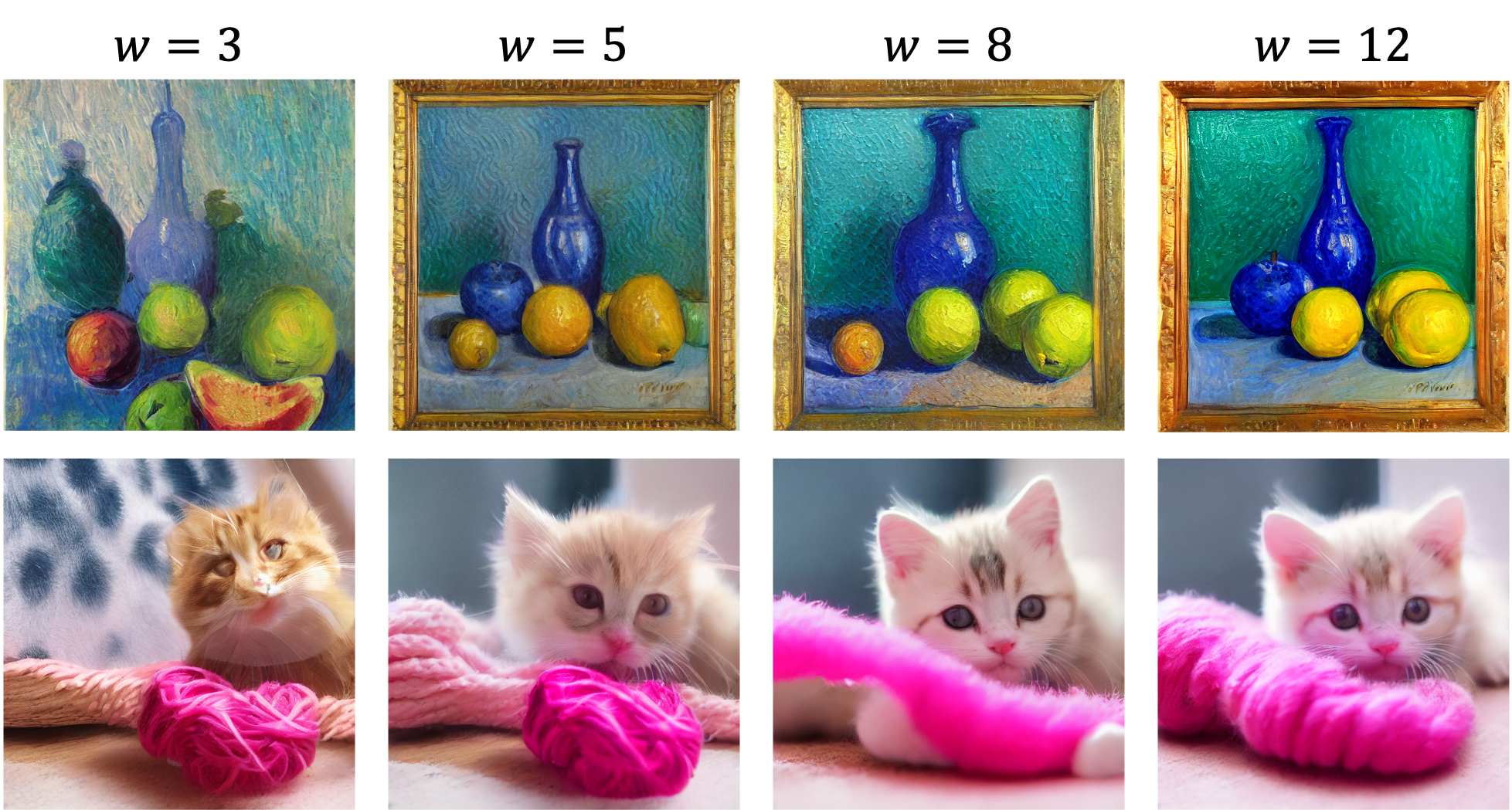}
    \caption{}
    \label{subfig:image1}
  \end{subfigure}
  \hspace{0.1cm}
  \raisebox{0.095\height}{\rule{0.5pt}{0.207\textheight}}
  \begin{subfigure}{.33\textwidth}
    \centering
    \includegraphics[width=0.92\linewidth]{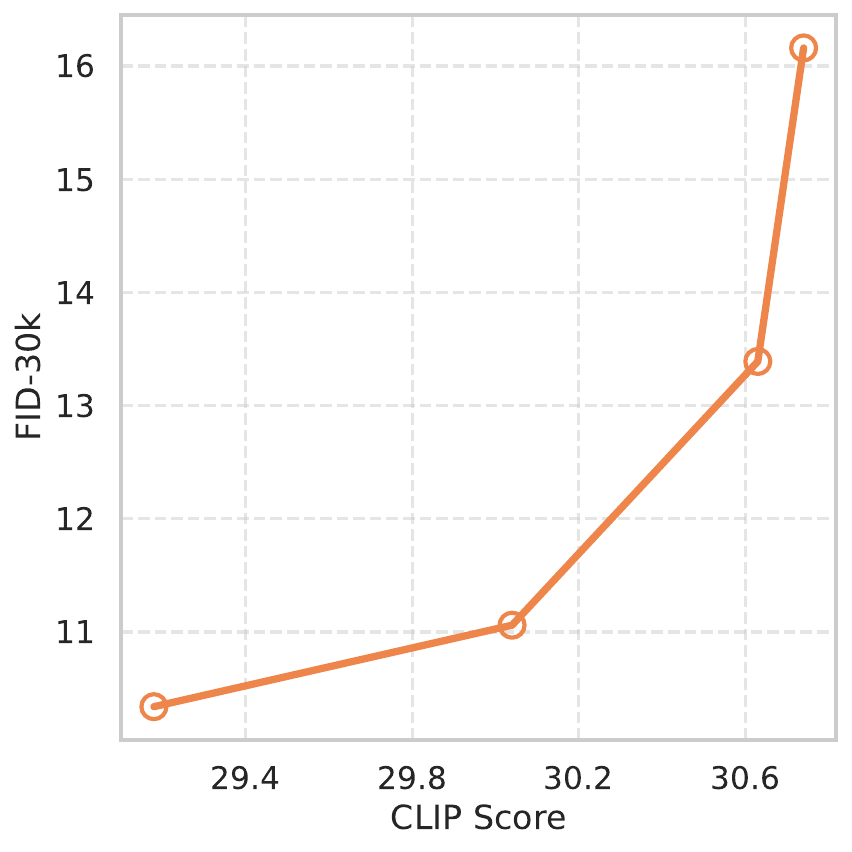}
    \caption{}
    \label{subfig:image2}
  \end{subfigure}

  \caption{(a) Visualization for different guidance scale $w$ on SPLAM. (b) The trade-off curve of applying difference guidance scale. $w$ increases from $\{3.0, 5.0, 8.0, 12.0\}$.}
  \label{fig.AblationGuidanceScale}
  \vspace{-2mm}
\end{figure}
\noindent\textbf{Skipping Step Size \& Training Cost}
\cref{subfig:stepsize} ablates the skipping step size during training, where we compare SPLAM with or without the multiple estimation strategy (\cref{subsec:sld}) and LCM.
We can observe that:
1) Without multiple estimation, when the skipping step size $k$ is increasing, LCM suffers a more drastic decline in performance due to heightened optimization challenges for sub-path learning.
Through leveraging our proposed Sub-Path Linear ODE, SPLAM can progressively learn the $dist_\Delta$ and effectively alleviate this collapse.
2) Equipped with the multiple estimation strategy, SPLAM is capable of stably maintaining high image fidelity with large steps.
Moreover, we compare the convergence trends between our method and LCM during training, as depicted in \cref{subfig:iteration}.
When $k=20$, although our metrics initially converge more slowly during the early stages, the performance of our method gradually surpasses LCM by a large margin.
It indicates that our training strategy provides a more effective learning objective, enabling SPLAM to achieve a better result, while LCM quickly becomes overfitted.
As $k$ raised to 100, larger skipping step size brings SPLAM faster convergence that needs just 2K to 6K iterations which requires about only 6 A100 GPU days training, facilitating practical applications with fewer resources.
Note that LCM needs 10k+ iterations for optimal performance which costs about 16 A100 GPU days and can not be applied to larger skipping step size due to the serious performance gap.
\\
\noindent\textbf{Approximated Error Estimation for SPLAM.}
To illustrate the efficacy of our approach, we directly estimate the denoising mapping error between two adjacent samples on the PF-ODE:
$\delta(t,k)=\mathbb{E}[|\bm{f}_{\bm{\theta}}(\bm{x}_{t_{n+k}, t_{n+k}}), \bm{f}_{\bm{\theta}}(\bm{x}_{t_n}, t_n))|]$, which is firstly defined in \cref{eq.CDLoss}.
The results are shown in \cref{subfig:diff}.
We randomly selected 1000 samples from the COCO dataset and simulated adjacent points on the ODE by adding the same noise with adjacent timesteps.
We utilize $k=20$ and the corresponding 50 timesteps for the DDIM scheduler, disregarding steps smaller than 100 due to their relatively larger simulation deviation.
It can be seen that, especially at larger timesteps, the error $\delta$ of our SPLAM is further reduced (about $10\%$ at $t=800$).
This observation substantiates that SPLAM indeed contributes to minimizing approximated errors, boosting the model's capacity for high-quality image generation.

\noindent\textbf{The Effect of Guidance Scale $w$.}
The guidance scale $w$ is a critical hyper-parameter in Stable Diffusion~\cite{LDM,CFG}, with its adjustment allowing users to alter the semantic alignment and the quality of the generated image.
In this study, we also examine the impact of varying the guidance scale $w$ for our SPLAM based on SDv1.5, which is visualized in \cref{fig.AblationGuidanceScale}.
As well as vanilla Stable Diffusion, while a higher $w$ value contributes to better sample quality as reflected by CLIP Scores, it concurrently leads to a degradation in FID performance and oversaturation.

\begin{figure}[t]
    \centering
    \includegraphics[width=0.95\textwidth]{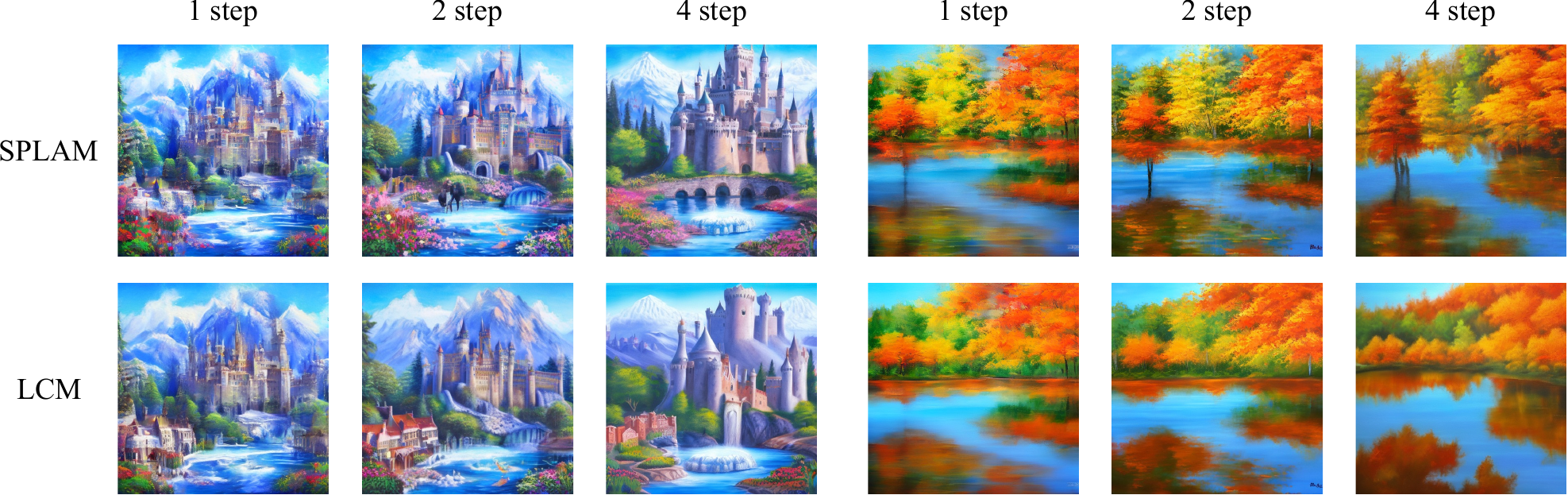}
    \caption{Comparsion of our SPLAM and LCM~\cite{LCM} in 1,2 and 4-step generation. The results of LCM are based on our reproduction as illustrated in \cref{sec:exp1}. SPLAM has generated consistently higher-quality images that are clearer and more detailed. Noteworthy is the remarkable performance of SPLAM in the 2-step generation, which aligns closely with the 4-step generation results of LCM, highlighting the efficiency and effectiveness of our approach in producing high-fidelity images with fewer generation steps.}
    \label{fig.ImgComp}
    \vspace{-2mm}
\end{figure}
\label{sec:exp3}
\begin{figure}[!tb]
    \begin{subfigure}[!t]{\textwidth}
        \centering
        \includegraphics[width=\textwidth]{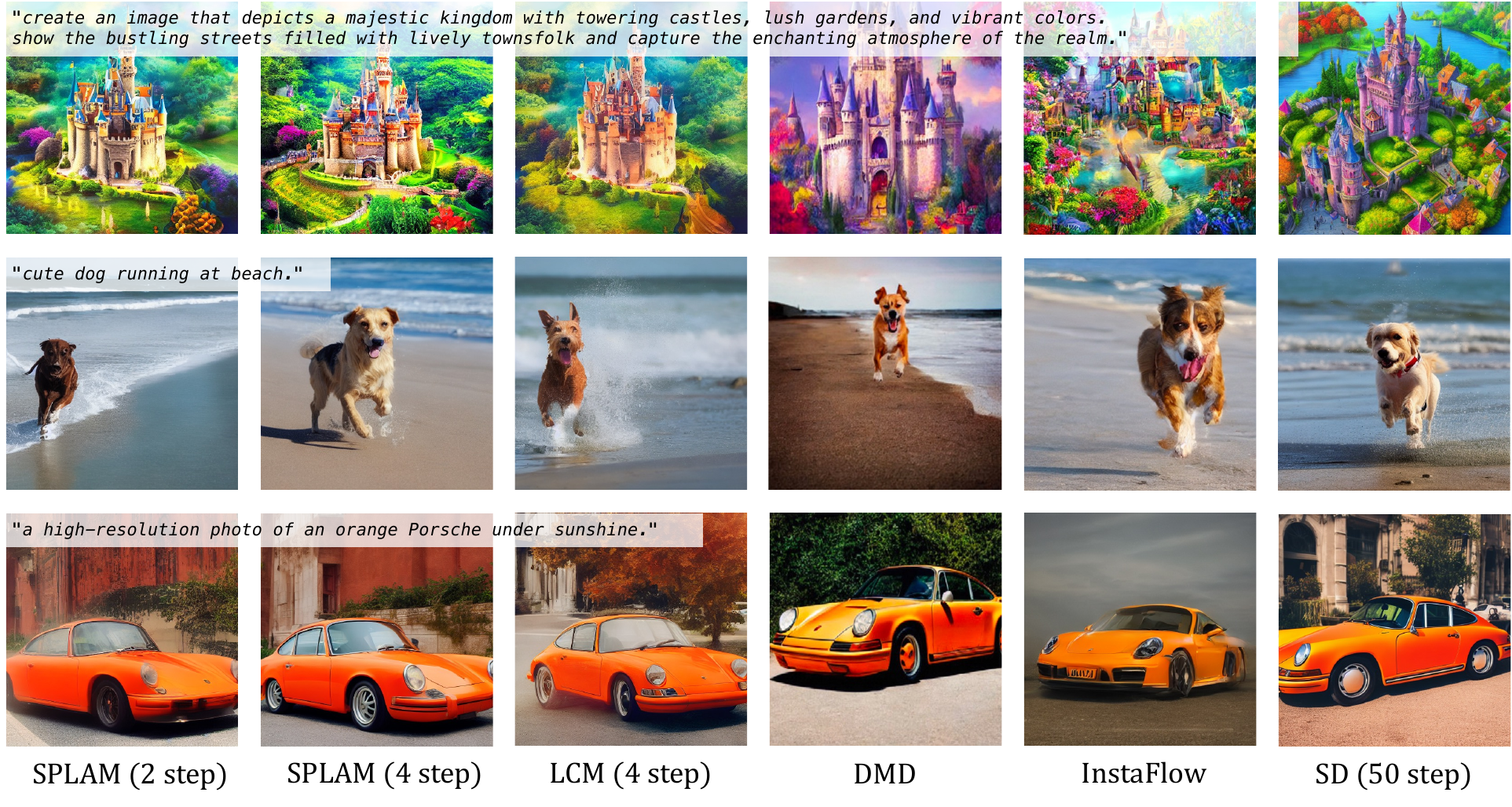}
        \caption{}
        \label{fig.Qresults1}
    \end{subfigure}
    \begin{subfigure}[t]{\textwidth}
    \centering
    \includegraphics[width=\textwidth]{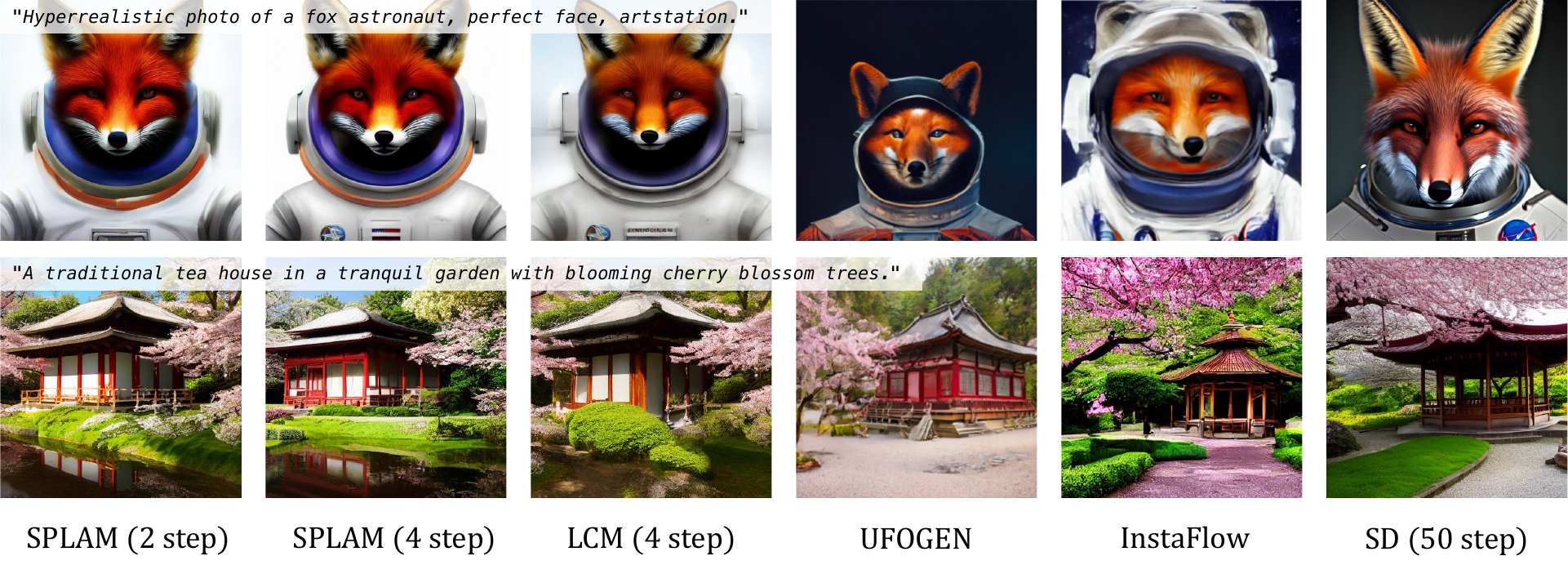}
    \caption{}
    \label{fig.Qresults2}
    \end{subfigure}
    \caption{Qualitative Results. The text prompts are selected from DMD~\cite{dmd} in (a) and UFOGEN~\cite{xu2023ufogen} in (b), and the results of the two are also cited from respective papers. Clearly, SPLAM demonstrates the best generation quality in 4-step generation except for the SD models. When decreasing the sampling step to 2, SPLAM still maintains a comparable performance, which generates even better results than 4-step LCM~\cite{LCM}.}
    \label{fig.Qresults}
    \vspace{-2mm}
\end{figure}

\subsection{Qualitative Results} 
\label{sec:exp3}
To emphasize the boosted generation quality of our SPLAM, we display the 1,2 and 4-step generation results with the comparison to LCM~\cite{LCM} in \cref{fig.ImgComp}. 
Moreover, we compare our SPLAM distilled from SDv1.5~\cite{LDM} with the most advanced accelerating diffusion models in \cref{fig.Qresults}, which demonstrate that our SPLAM has achieved the best generation quality across the existing methods.
\section{Conclusion}
In this paper, we propose a novel approach Sub-Path Linear Approximation Models (SPLAM) for accelerating diffusion models. SPLAM leverages the approximation strategy in consistency models and considers the PF-ODE trajectories as a series of interconnected sub-paths delineated by sampled points. Guided by the optimization direction charted by each sub-path, Sub-Path Linear (SL) ODEs also enable our approach 
to progressively and continuously optimize the approximated learning objectives and thus construct the denoising mappings with smaller cumulative errors.
We also develop an efficient distillation procedure for SPLAM to enable the incorporation of latent diffusion models. Extensive experiments on LAION, MS COCO 2014 and MS COCO 2017 datasets have consistently demonstrated the superiority of our method across existing accelerating diffusion approaches in a few-step generation with a fast training convergence. 

%\clearpage  % TODO REVIEW/FINAL: This \clearpage needs to be removed from both review and camera-ready versions.

\section*{Acknowledgments}
This work is supported by the National Key R$\&$D Program of China (No. 2022ZD0160900), the National Natural Science Foundation of China (No. 62076119, No. 61921006), the Fundamental Research Funds for the Central Universities (No. 020214380119), and the Collaborative Innovation Center of Novel Software Technology and Industrialization.

\vspace{1cm}

\appendix

\noindent\textbf{\Large{Appendix}}

\section{Implementation Details}
\noindent\textbf{Common Settings:} On text-to-image generation task, we train two models with pre-trained Stable Diffusion-V1.5 (SDv1.5) and Stable Diffusion-V2.1-base (SDv2.1-base) as teacher models respectively.
Following the setting of \cite{LCM}, the training dataset is one subset of LAION-5B\cite{schuhmann2022laion}: LAION-Aesthetics-6+, which comprises 12M text-image pairs with predicted aesthetics scores higher than 6.
We choose DDIM-Solver \cite{DDIM} as the ODE solver $\Phi$ and uniformly set the solver skipping step size $k_\phi$ to 20 across all experiments. The training process is optimized using the AdamW optimizer \cite{loshchilov2017decoupled}, with a learning rate configured to 8e-6 and a weight decay of 0. To ensure the stability of gradient updates, we implement gradient clipping with an L2 norm threshold of 10, and apply Exponential Moving Average (EMA) with a decay rate of 0.95. The models are trained with a total batch size of 1024 distributed across 16 A100 GPUs consistently.
For evaluation, we adopt the common practice that uses Fréchet Inception Distance (FID) which measures image quality, and CLIP Score~\cite{radford2021learning} which measures text-to-image alignment, which we employ LAION OpenCLIP ViT-G/14 \cite{openclip} to calculate. For LAION, as stated in \cite{LCM}, we generate 30k images from 10k text prompts as the evaluation set. For MSCOCO2014-30k (Zero-shot FID-30k) and MSCOCO 2017-5k, we follow the evaluation protocol described in \cite{imagen} and \cite{OnDistillation}, respectively.

\noindent\textbf{For SDv2.1-base:}
To make a fair comparison between LCM and our approach, we train our SPLAM with skipping step size $k=20$ and L2 distance metric function, which is consistent with \cite{LCM} (the multiple estimation strategy is disabled since $k\leq k_\phi$). The difference is that we trained our models with a fixed guidance scale $w=8$.
We also reproduce an SDv2.1-base LCM according to the training configuration outlined in \cite{LCM} while replacing the $w$-condition with the fixed guidance scale, which has also improved the performance.
For training costs, one GPU day can process 1.6K iterations with batch size of 1024 and without multiple estimation (ME) strategy.
We train SPLAM on SDv2.1-base with 80k iterations, which costs about 60 A100 GPU days.

\noindent\textbf{For SDv1.5:}
The guidance scale $w$ is set to $3$ to obtain the optimal FIDs, and we adopt the huber loss in~\cite{song2023improved} as we find that it provides faster convergence and better performance on two-step and one-step.
Guidance scale $w=8$ is also conducted and all ablation studies are under SDv1.5 with $w=8$ by default.
For training costs, we train SPLAM on SDv1.5 with multiple estimation (ME) described in Sec. 4.2 in the main paper, which executes solver $\Phi$ for $\frac{k}{k_\phi}$ times in one iteration. In our platform, when $k$=100 and $k_\phi=20$, one GPU day can process nearly half the amount of data compared to previously. As the skipping step size $k$ set to $100$ has shown fast convergence, it just costs 6k training iterations which require only 6 GPU days.

\section{Derivation of $\alpha(\gamma, t)$ and $\sigma(\gamma, t)$}

The perturbation process of DDPM \cite{DDPM} which corresponds to Variance Preserving (VP) SDE \cite{score-base}, could be given by the following Markov chain:
\begin{equation}
\label{eq:ddpm1}
    \bm{x}_t =  \frac{\alpha({t})}{\alpha({t - k})}\bm{x}_{t - k} + \sqrt{1 - \frac{\alpha(t)^2}{\alpha({t - k})^2}} \bm{\epsilon},
\end{equation}
where $\bm{\epsilon} \sim \mathcal{N}(\bm{0}, \bm{I})$.
For VP $\alpha(t)$ is the controlling schedule with $\alpha(0) = 1$, and $\sigma(t)$ is defined as $\sigma(t) = \sqrt{1 - \alpha({t})^2}$.

A nice property of the above process is that we can sample $\bm{x}_t$ at any arbitrary time step $t$ in a closed form:
\begin{equation}
\label{eq:ddpm2}
\begin{aligned}
    \bm{x}_{t}  & =\alpha({t}) \bm{x}_0 + \sigma(t) \bm{\epsilon} \\
                & = \alpha({t}) \bm{x}_0 + \sqrt{1 - \alpha({t})^2} \bm{\epsilon}.
\end{aligned}
\end{equation}

Accordingly, we have the posterior:
\begin{equation}
    p(\bm{x}_t | \bm{x}_0) = \mathcal{N}(\bm{x}_t; \alpha(t) \bm{x}_0, \sigma(t)^2 \bm{I})
\end{equation}

To obtain the posterior for $\bm{x}_{\gamma, t}$, where $p(\bm{x}_{\gamma, t}|\bm{x}_0) = \mathcal{N} (\bm{x}_{\gamma, t}; \alpha(\gamma, t) \bm{x}_0, \sigma(\gamma, t)^2 \bm{I})$,
we compute $\bm{x}_{\gamma, t}$ as described in Eq. (12) in the main paper, and substitute \cref{eq:ddpm1} and \cref{eq:ddpm2} into it:
\begin{equation}
\begin{aligned}
    \bm{x}_{\gamma, t} &= (1 - \gamma) \frac{\alpha(t)}{\alpha(t - k)} \bm{x}_{t - k} + \gamma \bm{x}_{t} \\
                       &= (1 - \gamma) \frac{\alpha(t)}{\alpha(t - k)} \bm{x}_{t - k} + \gamma \left(\frac{\alpha(t)}{\alpha(t - k)} \bm{x}_{t - k} + \sqrt{1 - \frac{\alpha(t)^2}{\alpha(t - k)^2}} \bm{\epsilon}\right) \\
                       &= \frac{\alpha(t)}{\alpha(t - k)} \bm{x}_{t - k} + \gamma \sqrt{1 - \frac{\alpha(t)^2}{\alpha({t - k})^2}} \bm{\epsilon},
\end{aligned}
\end{equation}

Thus $\bm{x}_{\gamma, t}$ could be represented as a merging of two \textit{independent} variables. Since we have  $p_t(\bm{x}_{t-k}|x_0)=\mathcal{N}(\bm{x}_{t-k};\alpha(t-k)\bm{x}_0,\sigma(t-k)^2\bm{I})$, the closed-form expression for the mean and variance in $p_t(\bm{x}_{\gamma, t}|\bm{x}_0)$ should be:
\begin{equation}
\begin{aligned}
    \alpha(\gamma, t) &= \frac{\alpha(t)}{\alpha(t - k)} * \alpha(t - k) + 0 = \alpha_{t}, \\
    \sigma(\gamma, t)^2 &= \frac{\alpha(t)^2}{\alpha(t - k)^2} \sigma(t - k)^2 + \gamma^2 \left( 1 - \frac{\alpha(t)^2}{\alpha(t - k)^2}  \right),
\end{aligned}
\end{equation}
where $\sigma(t) = \sqrt{1 - \alpha(t)^2}$ as defined in this noise schedule.

Subsequently, we can compute the error $e^{\text{VP}}_\sigma$ between the derived result and the empirical value $\sigma'(\gamma, t)=(1-\gamma)*\frac{\alpha(t)}{\alpha(t-k)}\sigma(t-k)+\gamma*\alpha(t)$ that we have employed in the main paper:
\begin{equation}
\begin{aligned}
    e^{\text{VP}}_{\sigma}=    &\sigma(\gamma, t)^2 - \sigma'(\gamma, t)^{2} \\
    =   &\frac{\alpha(t)^2}{\alpha(t - k)^2} (1 - \alpha_{t - k}^2) + \gamma^2 \left( 1 - \frac{\alpha(t)^2}{\alpha(t - k)^2} \right) - \left( \gamma \sigma_t + (1 - \gamma) \frac{\alpha(t)}{\alpha(t - k)} \sigma_{t - k} \right)^2 \\
    =   &2 \gamma (1 - \gamma) \frac{\alpha(t)}{\alpha(t - k)} \sigma(t - k) \left( \frac{\alpha(t)}{\alpha(t - k)} \sigma(t - k) - \sigma(t) \right)\\
    = & 2 \gamma (1 - \gamma) \frac{\alpha(t)}{\alpha(t - k)} \sigma(t - k) \left(\sqrt{\frac{\alpha(t)^2}{\alpha(t-k)^2}-\alpha(t)^2}-\sqrt{1-\alpha(t)^2}\right).
\end{aligned}
\label{eq.ErrorVP}
\end{equation}

As denoted in \cref{eq.ErrorVP}, the estimated error $e^{\text{VP}}_\sigma\equiv 0$ for any endpoints on the PF-ODE trajectories, specifically when $\gamma = 1 $ or $0$, and thus ensures the fidelity for generation. It is observed that as the value of $k$ increases, a concomitant rise in error is typically expected. However, in our experimental analysis, we discovered that even for the scenario where $k=100$, training with $\sigma(\gamma,t)$ or $\sigma'(\gamma,t)$ yields nearly the same results. This indicates that the training on SL-ODEs could tolerate such small errors in the intermediate sampled points.

\section{Strategies for Constructing Sub-path ODE}
To construct approximated paths for the sup-paths on the PF-ODE trajectories, another intuitive way except for our SL ODEs is to directly connect the corresponding endpoints $(\bm{x}_{t-k},t-k)$ and $(\bm{x}_t,t)$, and we call it as \textit{Direct Linking} (DL) ODE. The DL-ODE thus can be formulated as:
\begin{equation}
    \begin{aligned}
    &\bm{x}_{\gamma,t}= \bm{x}_{t-k}+\gamma(\bm{x}_t-\bm{x}_{t-k})\\
    &d\bm{x}= [\gamma*(\bm{x}_t-\bm{x}_{t-k})]d\gamma.\\
    \end{aligned}
    \label{eq.DLODE}
\end{equation}
And the approximated learning objective for DL ODEs is:
\begin{equation}
    \small
    \mathcal{L}_{DL}(\bm{\theta},k)= \mathbb{E}[|\frac{\bm{x}_{\gamma,t}-\sigma_{DL}(t,\gamma)\bm{\epsilon_\theta}(\bm{x}_{\gamma,t},\gamma,t)}{\alpha_{DL}(\gamma,t)}-\frac{\bm{x}_{t-k}-\sigma(t-k)\bm{\epsilon}_{\bm{\theta}}(\bm{x}_{t-k},t-k)}{\alpha(t-k)}|],
    \label{eq.DLObj}
\end{equation}
where the noise schedule $\alpha_{DL},\sigma_{DL}$ is defined by the marginal distribution of $\bm{x}_{\gamma,t}$: $p(\bm{x}_{\gamma,t}|x_0)= \mathcal{N}(\bm{x}_{\gamma,t};\alpha_{DL}(\gamma,t)\bm{x}_0,\sigma_{DL}^2(\gamma,t)\bm{I})$. We have derived the expression of $\alpha_{DL},\sigma_{DL}$, which should be :
\begin{equation}
    \begin{aligned}
    &\alpha_{DL}(\gamma,t) = \gamma*\alpha(t)+(1-\gamma)*\alpha(t-k)\\
    &\sigma_{DL}(\gamma,t)^2 = (1-\gamma+\frac{\alpha(t)}{\alpha(t-k)}*\gamma)^2*\sigma(t-k)^2 + \gamma^2 \left( 1 - \frac{\alpha(t)^2}{\alpha(t - k)^2}\right).\\
    \end{aligned}
    \label{eq.DLSchedule}
\end{equation}
Comparing \cref{eq.DLSchedule} and Eq. (15) in the main paper, the learning on SPLAM could separate the optimization process for $dist_\Delta$ and $dist_{0,\theta}$, whereas the DL-ODE provides a blended estimation. We apply both strategies to the distillation process from SDv1.5, and denote the optimized models as Opt$_\text{DL}$ for the training on DL-ODEs and Opt$_\text{SL}$ for our SPLAM learning. The tested FIDs are shown in \cref{tab.DLAndSL}. While DL-ODE also makes a continuous  estimation for the sub-path learning objective and surely outperforms LCM, the entangled optimization for $dist_\Delta$ and $dist_{0,\theta}$ still restricts its optimized performance. In contrast, our approach consistently yields optimal FIDs, reaffirming its superiority in generating high-quality images.
\setlength{\tabcolsep}{12pt}
\begin{table}[t]
    \centering
    \caption{Comparison between the optimized performance based on DL ODEs and SL ODEs. We report the 4-step generation results for all methods with a guidance scale $w$=8.}
    \label{tab.DLAndSL}
    \begin{tabular}{l|cc}
    \toprule
    Methods       & FID-5k & FID-30k \\
    \midrule
    LCM~\cite{LCM} & 24.68 & 14.53 \\
    Opt$_{\text{DL}}$     & 24.20 & 13.89    \\
    Opt$_{\text{SL}}$ (SPLAM)    & \textbf{23.76} & \textbf{13.39}   \\
    \bottomrule
    \end{tabular}
\end{table}

\section{Additional Generated Results}

\subsection{Comparsion to LCM with pre-trained DreamShaper.}
\begin{figure}[tb]
    \centering
    \includegraphics[width=\textwidth]{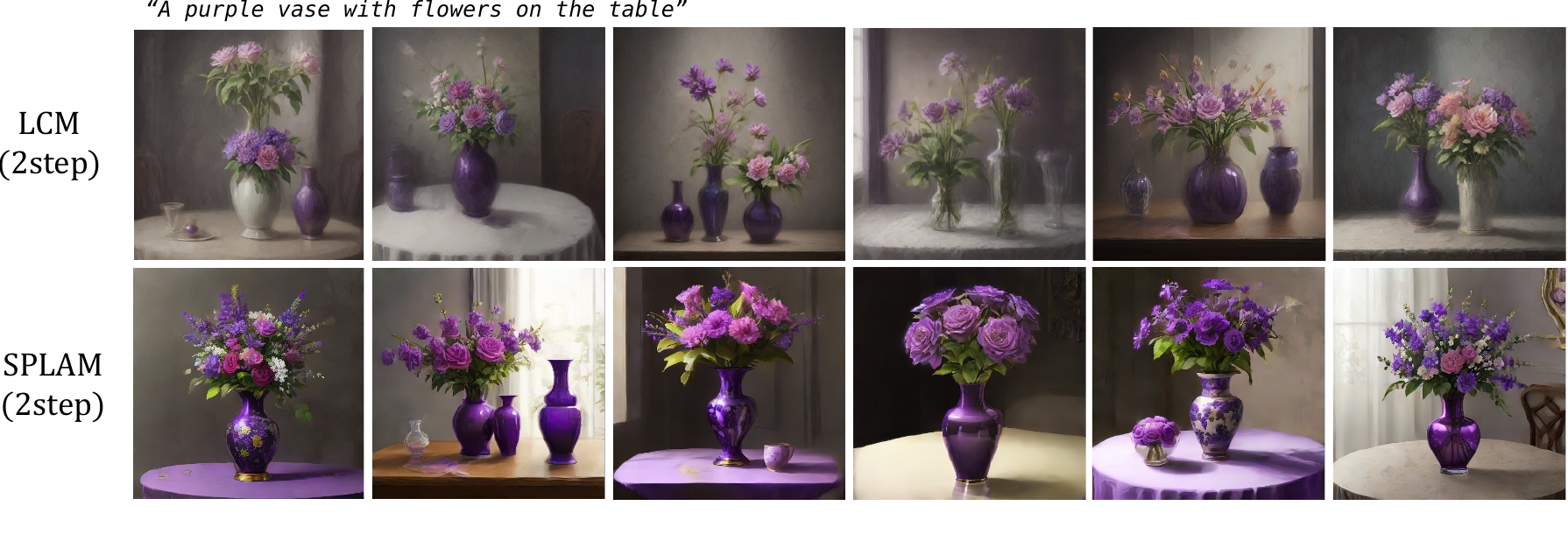}
    \includegraphics[width=\textwidth]{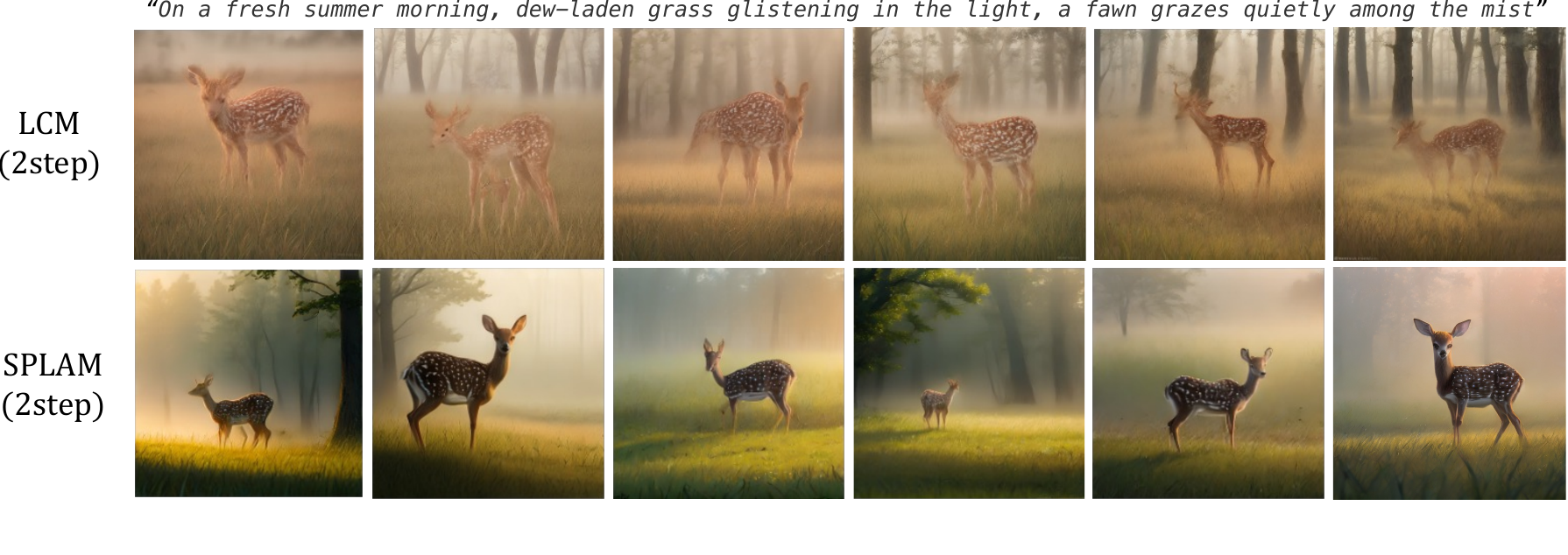}
    \includegraphics[width=\textwidth]{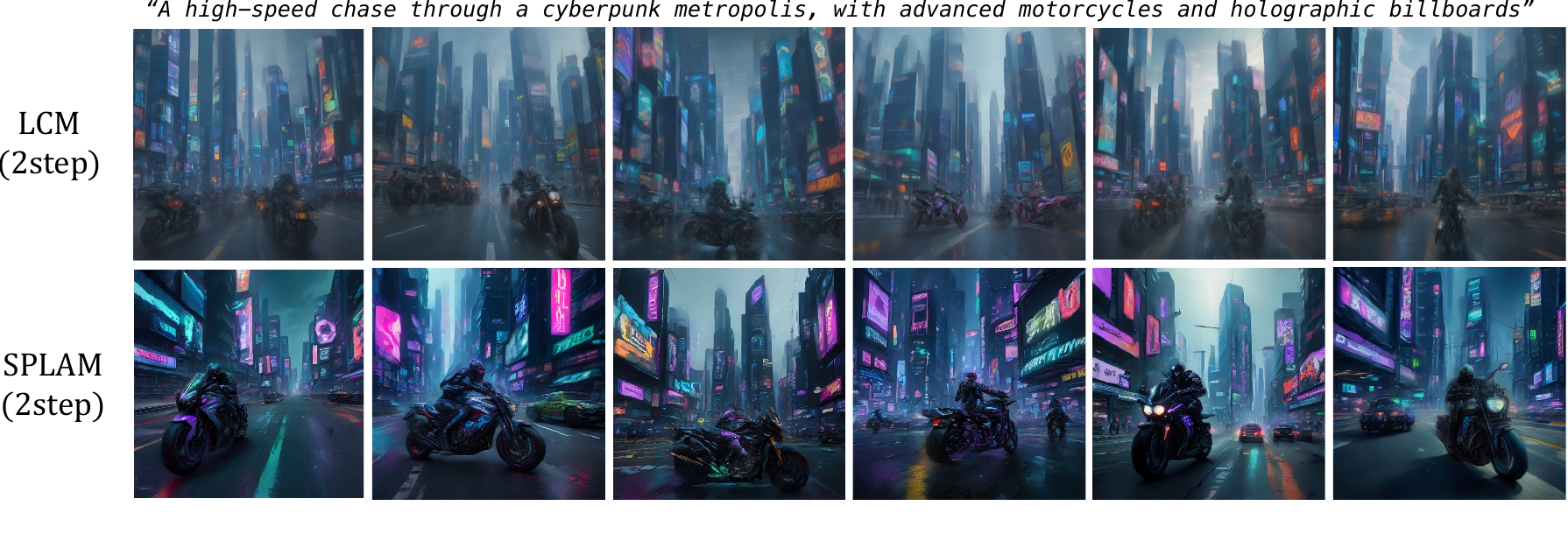}
    \caption{Generated samples on 2-step from our SPLAM and LCM distilled from DreamShaper-v7. We leverage the LCM-Dreamshaper-v7 checkpoint hosted on Hugging Face.}
    \label{fig:dreamshaper1}
\end{figure}

\begin{figure}[tb]
    \centering
    \includegraphics[width=\textwidth]{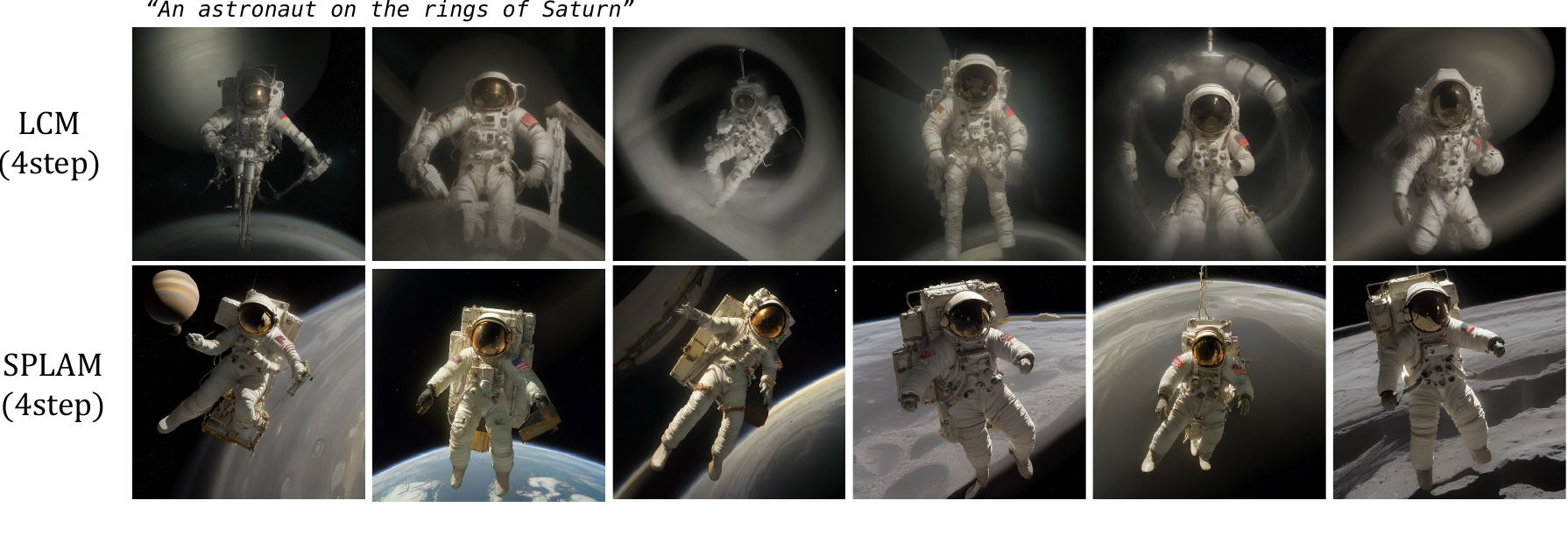}
    \includegraphics[width=\textwidth]{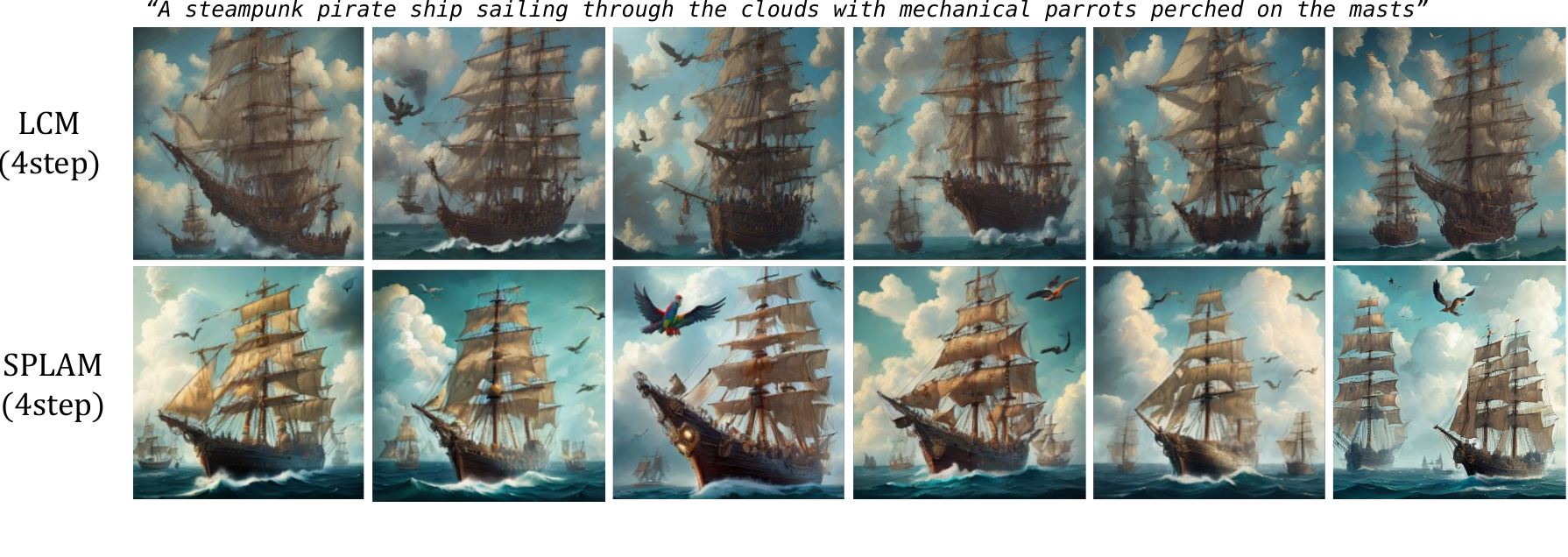}
    \includegraphics[width=\textwidth]{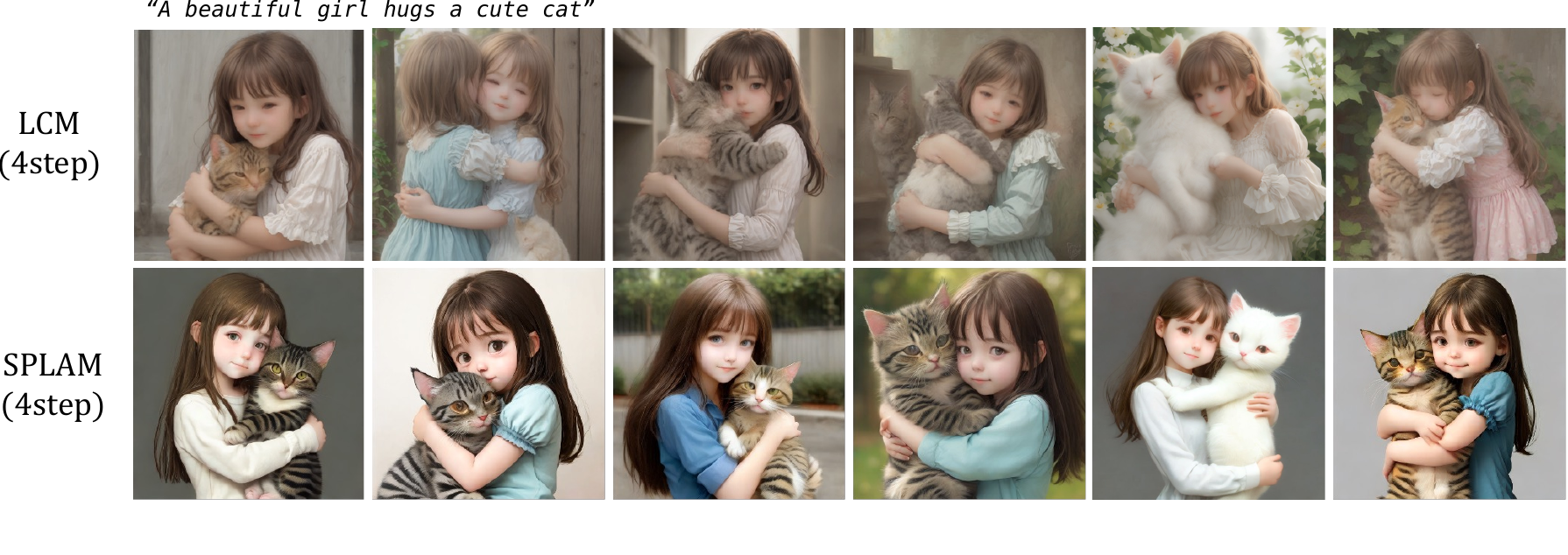}
    \caption{Generated samples on 4-step from our SPLAM and LCM distilled from DreamShaper-v7. We leverage the LCM-Dreamshaper-v7 checkpoint hosted on Hugging Face.}
    \label{fig:dreamshaper2}
\end{figure}

Since LCM has only released one version\footnote{\url{https://huggingface.co/SimianLuo/LCM_Dreamshaper_v7}} of SDv1.5 based model that is distilled from DreamShaper-v7, we train a SPLAM on DreamShaper-v7 as well and compare it to LCM, as shown in \cref{fig:dreamshaper1} and \cref{fig:dreamshaper2}.

\begin{figure}[!tb]
    \centering
    \includegraphics[width=\textwidth]{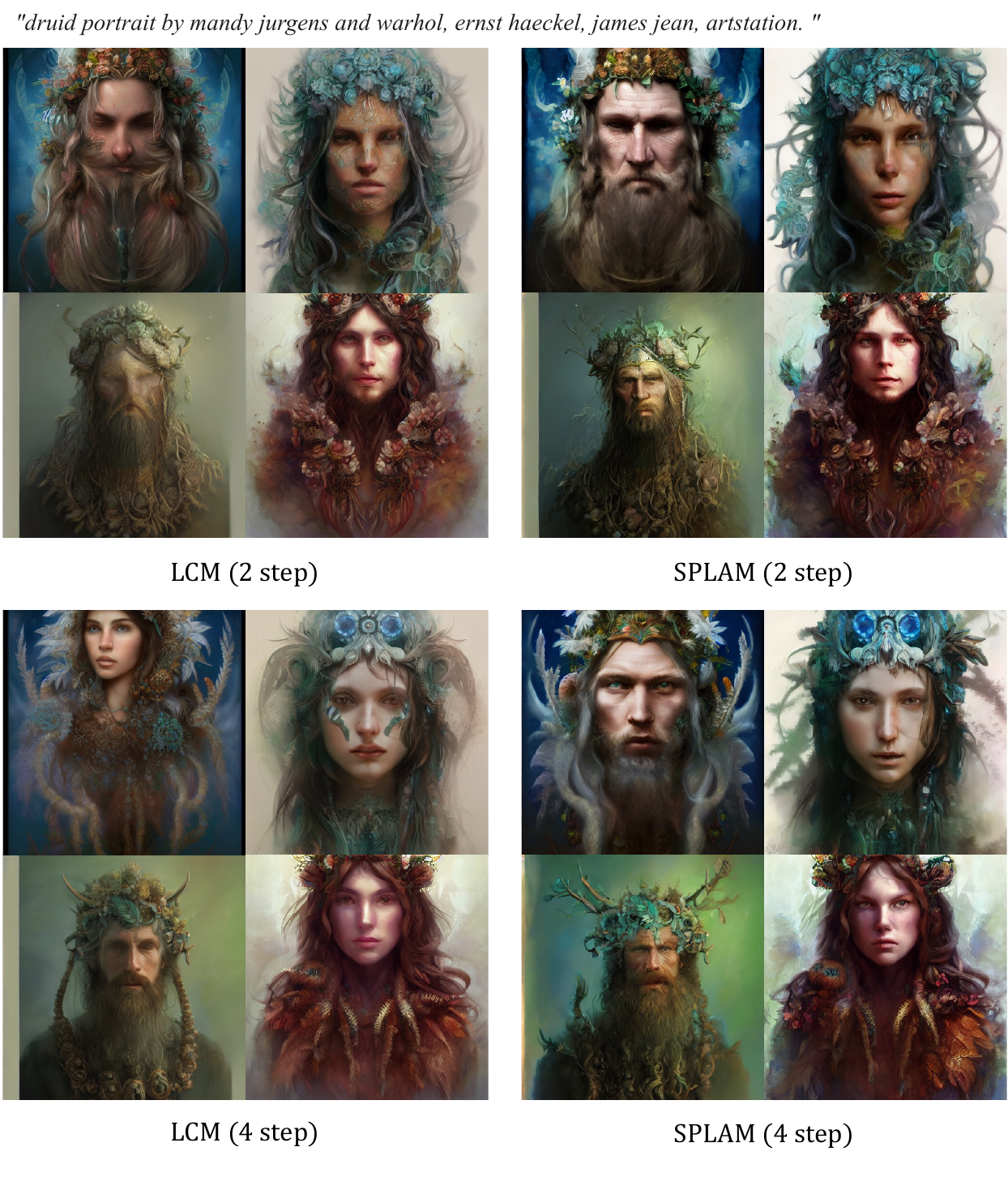}
    \caption{Generated samples from SPLAM (our method) and LCM.}
    \label{fig.Portrait}
\end{figure}

\begin{figure}[tb]
    \centering
    \includegraphics[width=\textwidth]{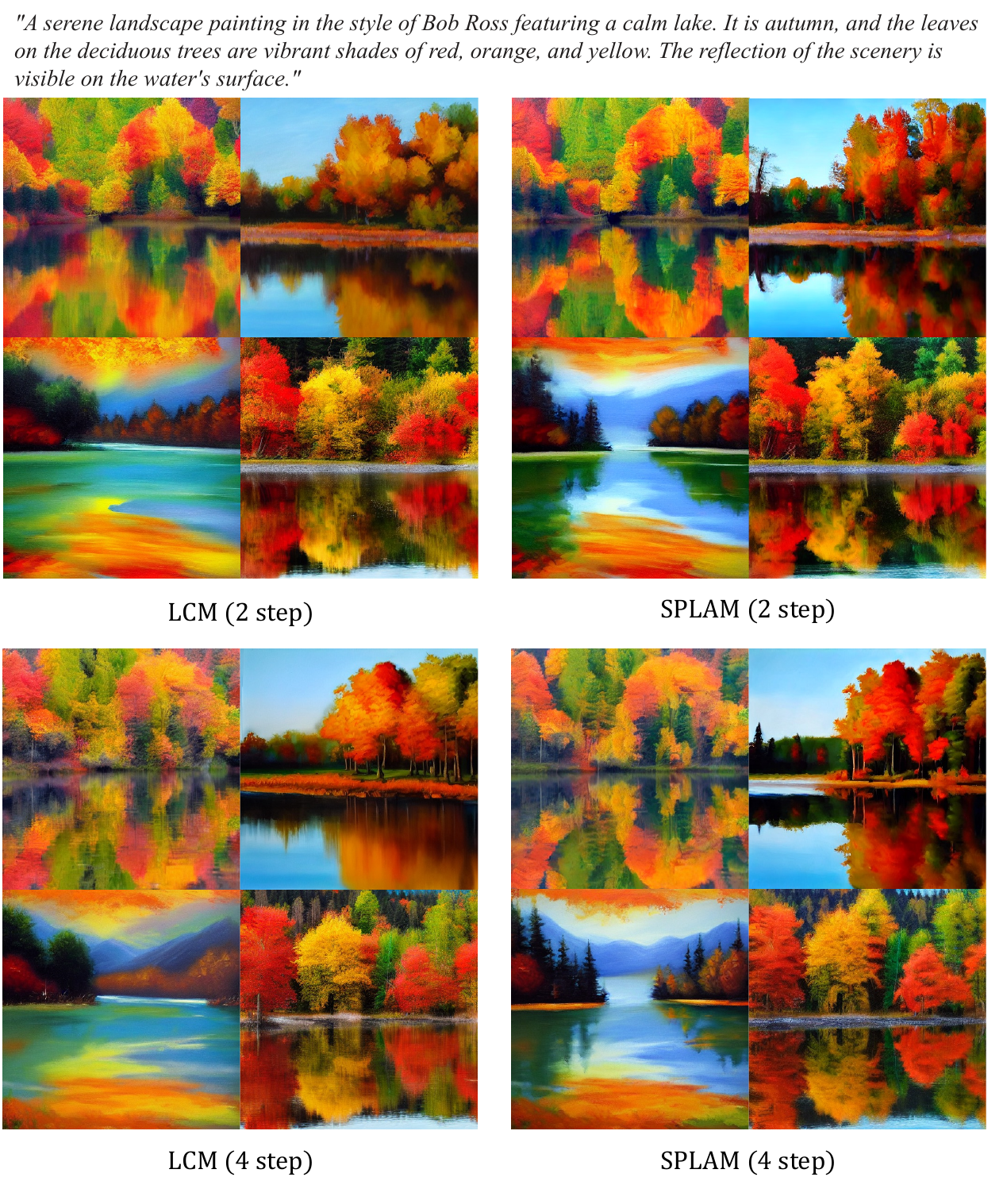}
    \caption{Generated samples from SPLAM (our method) and LCM.}
    \label{fig.Scene}
\end{figure}
\subsection{More Generated images with SD pre-trained Models.}
\begin{figure}[tb]
    \centering
    \includegraphics[width=\textwidth]{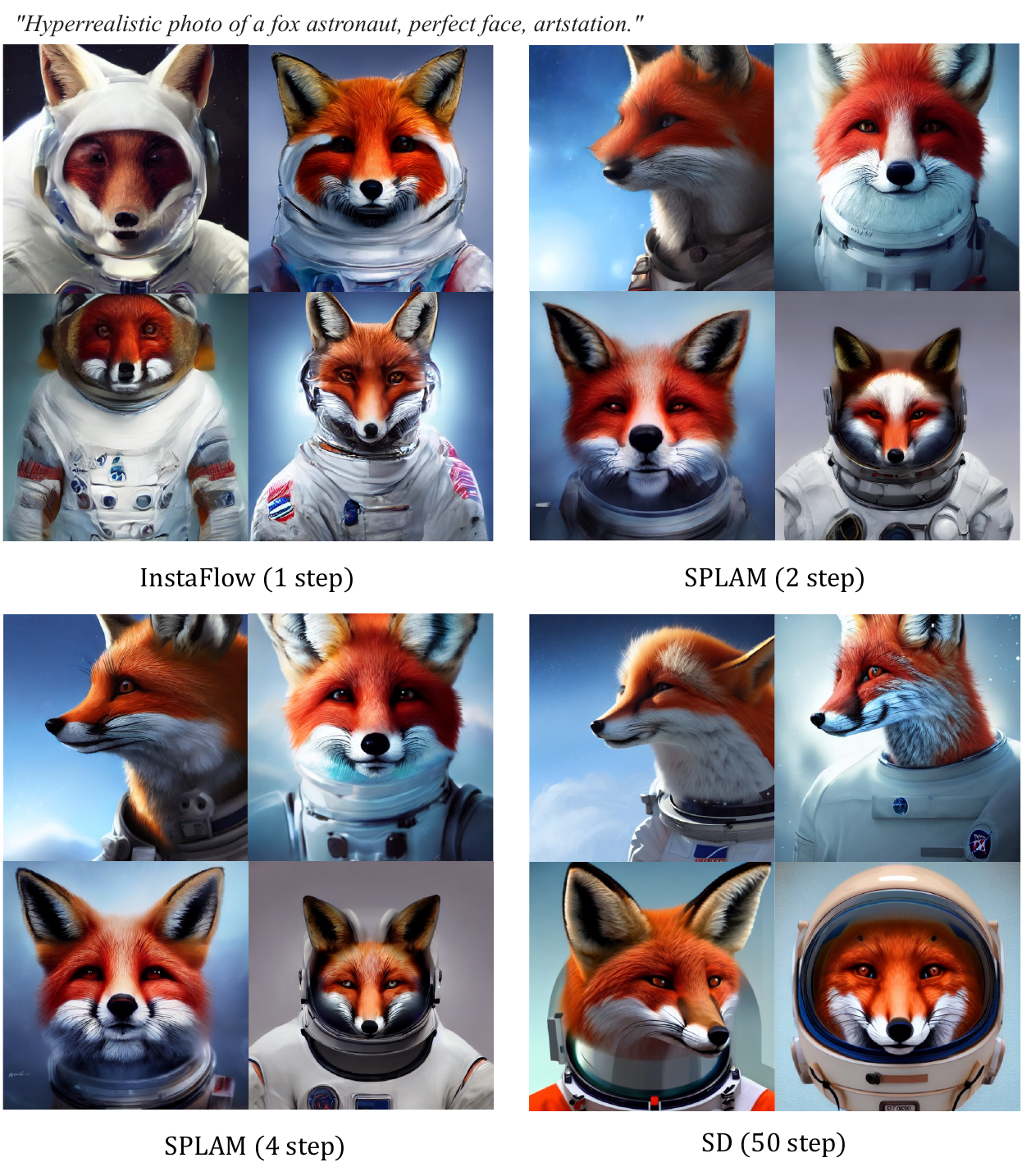}
    \caption{Comparison of our method SPLAM with one-step generation (InstaFlow~\cite{liu2023instaflow}) and Stable Diffusion~\cite{LDM}.}
    \label{fig.Fox}
\end{figure}

\begin{figure}[tb]
    \centering
    \includegraphics[width=\textwidth]{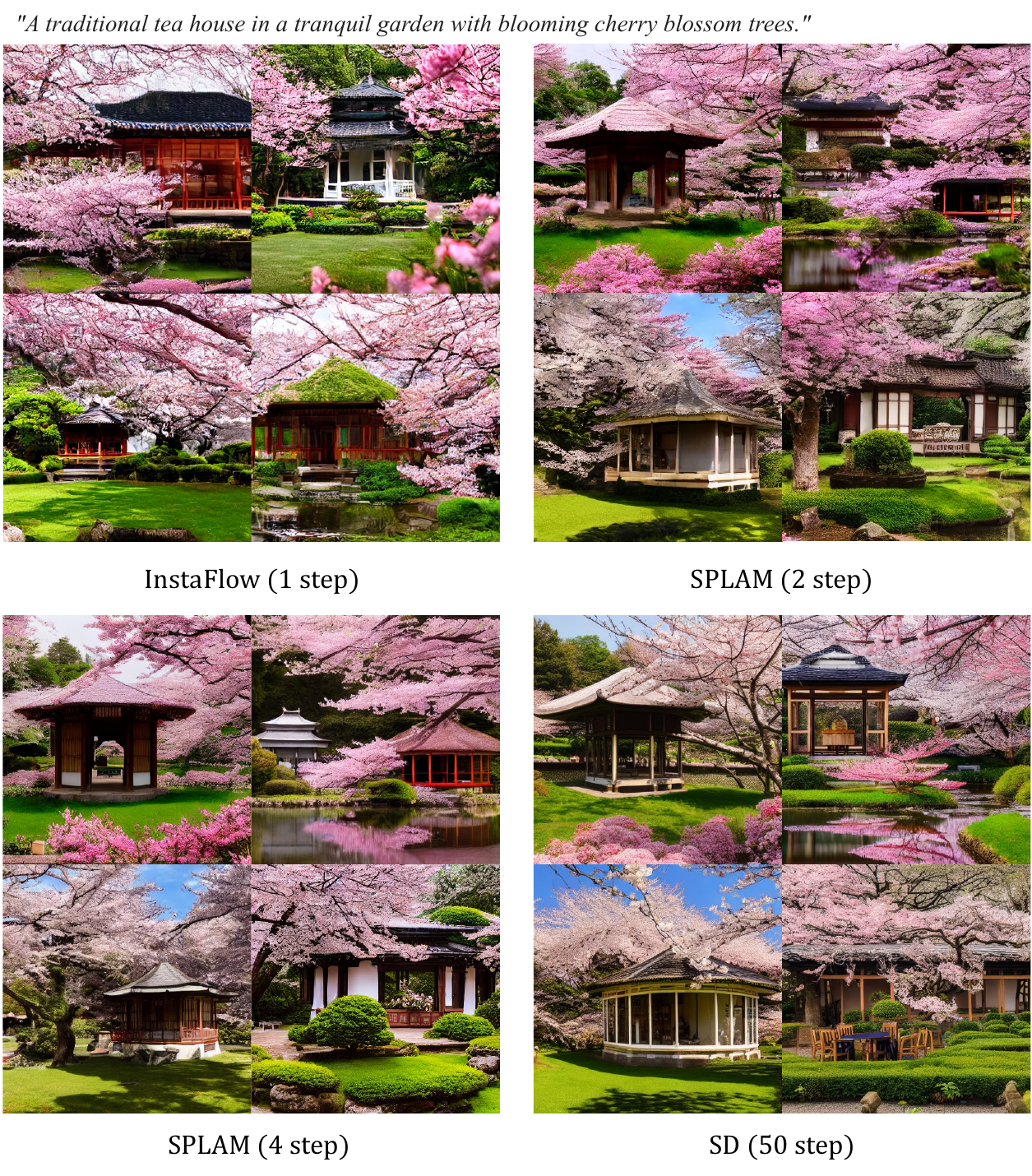}
    \caption{Comparison of our method SPLAM with one-step generation (InstaFlow~\cite{liu2023instaflow}) and Stable Diffusion~\cite{LDM}.}
    \label{fig.Teahouse}
\end{figure}

Here we provide more generated images from our SPLAM, which are compared with LCM~\cite{LCM}, the newly released one-step generation approach InstaFlow~\cite{liu2023instaflow} and Stable Diffusion~\cite{LDM}. Unless otherwise specified, our SPLAM's generated results are based on the model distilled from the open source SDv1.5, and the results for LCM are generated from our reproduction. The images are shown in \cref{fig.Portrait,fig.Scene,fig.Fox,fig.Teahouse}.

% ---- Bibliography ----
%
% BibTeX users should specify bibliography style 'splncs04'.
% References will then be sorted and formatted in the correct style.
%
\bibliographystyle{splncs04}
\bibliography{main}
\end{document}